\documentclass{article}

\usepackage[main,final,nonatbib]{neurips_2026}
\usepackage[numbers,sort&compress]{natbib}

\usepackage[utf8]{inputenc}
\usepackage[T1]{fontenc}
\usepackage{hyperref}
\usepackage{url}
\usepackage{booktabs}
\usepackage{amsfonts}
\usepackage{amsmath,amssymb}
\usepackage{amsthm}
\usepackage{nicefrac}
\usepackage{microtype}
\usepackage{xcolor}
\usepackage{graphicx}
\usepackage{wrapfig} 
\usepackage{titletoc}
\usepackage{pifont}
\newcommand{\gene}[1]{\textit{#1}}

\newtheorem{definition}{Definition}

\graphicspath{{figures/}}

\title{A low-rank decoder bottleneck bounds reliability in foundation-model perturbation prediction}

\author{%
  Zheyu Zhang$^{1}$ \quad Yuanhao Huang$^{2}$ \quad Fan Feng$^{2}$ \quad Jie Liu$^{1,2}$ \\
  $^{1}$Computer Science and Engineering Division \\
  $^{2}$Gilbert S. Omenn Department of Computational Medicine and Bioinformatics \\
  University of Michigan, Ann Arbor, MI 48109 \\
  \texttt{\{zheyuz, hyhao, fanfeng, drjieliu\}@umich.edu}
}

\begin{document}

\maketitle

\begin{abstract}
Foundation-model predictors of perturbation response are evaluated
primarily by aggregate accuracy. Using their predictions to prioritize
experiments requires assessing each prediction's reliability. We audit
the state-transition (ST) decoder used in STATE across three encoders
(SE-600M, scGPT, and Tahoe-x1) and four cell lines under CRISPRi. The
decoder's first attention layer has a low-rank output. Projecting this
output onto $4$--$32$ fixed directions estimated from training
activations preserves mean per-perturbation Pearson delta correlation
(PDC). This low-rank attention bottleneck bounds the decoder's
reliability by distance to training responses: PDC declines as the
cosine distance from a perturbation's observed response to its nearest
training response increases. We propose \emph{predicted-NN-distance},
the cosine distance from a predicted response to its nearest training
response, as a per-prediction confidence score. The score has a mean
absolute Spearman correlation of $0.81$ with PDC, close to the $0.88$
obtained with the observed response. Thresholds
calibrated on this score transfer across encoders and cell lines, and
the score remains predictive of PDC for drug perturbations.\footnote{Code is available at \url{https://github.com/HowIII/lowrank-decoder-reliability}.}
\end{abstract}

\section{Introduction}
\label{sec:intro}

\begin{figure}[t]
    \centering
    \includegraphics[width=\linewidth]{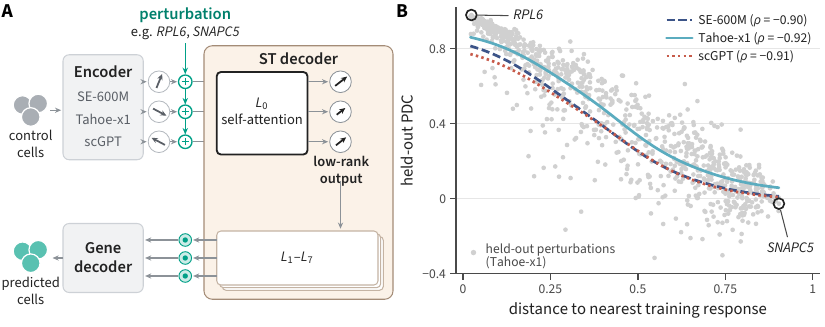}
\caption{\textbf{Per-prediction reliability of encoder$+$ST models
decays smoothly with gene-space distance to the training perturbations.}
\textbf{(A) Encoder coupled with ST decoder for perturbation
prediction.} An upstream encoder (any single-cell foundation model or
cell embedding, e.g., SE-600M, Tahoe-x1, scGPT) embeds each control
cell, and the perturbation embedding is added to every cell embedding
($\oplus$). The ST decoder ($8$ attention blocks) and a gene decoder
map these to predicted cells. Predictions are averaged into a
pseudobulk for evaluation (\S\ref{sec:methods}). Each cell is drawn as a
vector. Before $L_0$ (black outline) the cell vectors point in different
directions, and after it they point along one shared direction with
different lengths. This is the low-rank output of the $L_0$
self-attention (\S\ref{sec:results_mechanism}).
\textbf{(B) HepG2.} Each gray dot is one of the $945$ held-out
perturbations under Tahoe-x1$+$ST, plotted as Pearson delta
correlation (PDC, Eq.~\eqref{eq:pcc_delta}) against the cosine
distance in gene space from its response to the nearest training
response. Lines are kernel-smoothed trends per encoder (Spearman
$\rho$ in the legend). \gene{RPL6} (small distance) is predicted
reliably and \gene{SNAPC5} (large distance) is not. The pattern
replicates across all four cell lines
(\S\ref{sec:supp_trust_boundary_4cells}). The deployable analog
(Definition~\ref{def:pred_nn_dist}) and the calibrated threshold are
developed in \S\ref{sec:results_deployable}.}
\label{fig:family_interpolation}
\end{figure}

Predicting how cells respond to a genetic or chemical intervention
would let biologists choose which perturbations to test, a goal often
described as building a \emph{virtual
cell}~\citep{bunne2024build,rood2024toward}.
Observational
single-cell atlases record correlations across millions of
cells~\citep{regev2017human,tarhan2023single}, and CRISPR-based
Perturb-seq~\citep{dixit2016perturb,replogle2022mapping,nadig2024transcriptome}
measures the effect of intervening directly. 
The perturbation space, however, spans
tens of thousands of genes, combinatorial pairs, and diverse cellular
contexts, far more than any lab can screen. Predictive models are
therefore needed to extrapolate from measured to unmeasured perturbations.
Single-cell foundation
models~\citep{theodoris2023transfer,cui2024scgpt,hao2024large,pearce2025cross,tahoex1_2025}
pretrain encoders on large single-cell atlases, and models trained on
perturbation screens predict responses from the encoders'
representations. STATE~\citep{adduri2025predicting} pairs a cell-state
encoder with a trainable state-transition (ST) decoder: the encoder
represents control cells, and the decoder combines these representations
with a perturbation embedding to predict post-perturbation gene
expression (Figure~\ref{fig:family_interpolation}A). Later models pair
this decoder with other encoders or use in-context conditioning or
generative diffusion
instead~\citep{zhang2025tahoe,cheng2026harnessing,taggart2026gpl,wang2026xcell,yuan2026perturbdiff,he2026squidiff,dong2026stack}.

Recent benchmarks find that simple linear baselines can match these
models on aggregate
metrics~\citep{ahlmann2025perturbation,csendes2025benchmarking,wenteler2024perteval,wong2025simple},
and \citet{vinastorne2026systema} show that a predictor can score well
on these metrics by capturing response patterns shared across
perturbations. An experiment, however, tests a specific intervention in
a specific cellular context, usually one outside the training data. We
ask why simple baselines can match these models, and how a user can tell
whether a given prediction can be trusted.

Consider two CRISPRi knockdowns in HepG2, a human liver cell line, both
held out from training. With each of three foundation-model encoders, the
ST decoder predicts the response to \gene{RPL6} accurately and the
response to \gene{SNAPC5} poorly. \gene{RPL6} encodes a ribosomal
protein, and knockdowns of other ribosomal genes in the training data
produce similar responses. The \gene{SNAPC5} response lies far from all
training responses. A user who sees only the two predictions needs a
way to tell which one to act on.
The contrast holds across held-out perturbations. We measure accuracy by
per-perturbation Pearson delta correlation (PDC;
Eq.~\eqref{eq:pcc_delta}), the correlation across genes between predicted
and observed expression changes relative to control. PDC declines smoothly
as the cosine distance from an observed response to its nearest training
response increases (Figure~\ref{fig:family_interpolation}B).

The three encoders (SE-600M~\citep{adduri2025predicting},
scGPT~\citep{cui2024scgpt}, and Tahoe-x1~\citep{tahoex1_2025}) use
different pretraining recipes, yet the decline has the same shape with
each. We therefore audit the component they share, the ST decoder, across
four CRISPRi cell lines (\S\ref{sec:methods}). Its first attention layer
is a low-rank bottleneck: for each perturbation, the layer adds nearly
the same vector to every control cell. Interventions on this vector with
SE-600M and Tahoe-x1 show that the decoder tells perturbations apart by
the vector's direction. Projecting the layer's output onto $4$--$32$
directions estimated from training activations preserves PDC, so
what the later layers need from this layer lies in a few directions
fixed by the training perturbations. A perturbation whose response needs
other directions is predicted poorly. A linear read-out of the decoder's
inputs is nearly as accurate as the full decoder, and an oracle that
averages the training responses nearest to a perturbation's observed
response is more accurate. The decoder's accuracy therefore comes from
the similarity of a held-out response to training responses, and this
is why simple baselines can match it.

We propose \emph{predicted-NN-distance}
(Definition~\ref{def:pred_nn_dist}), the cosine distance from the
predicted response to its nearest training response, as a per-prediction
confidence score. It needs only the prediction and the training
responses, and its correlation with PDC is nearly as high as that of the
distance computed from the observed response (Spearman $|\rho|\approx0.81$
against $0.88$).
A threshold on the score fitted on the other encoders and cell lines
selects accurate predictions on a held-out one without recalibration.

\vskip 0pt plus 1fil\penalty0\vskip 0pt plus -1fil
Our main contributions are:

\begingroup\makeatletter\@beginparpenalty=10000\makeatother
\begin{itemize}
\setlength{\itemsep}{2pt}

\item We identify a low-rank bottleneck in the ST decoder: its first
  attention layer passes each perturbation to the later layers through a low-rank output, and its directions lie in a small subspace fixed by the training perturbations (\S\ref{sec:results_mechanism}).

\item We show that this bottleneck explains why simple baselines match
  these models: accuracy falls with distance from the nearest training
  response, and on perturbations far from training the decoders are no
  more accurate than a linear model on their inputs
  (\S\ref{sec:results_trust_boundary},~\ref{sec:results_mechanism}).

\item We propose predicted-NN-distance, a confidence score for each
  prediction computed from the prediction and the training data. One
  threshold on this score selects accurate predictions across encoders and
  cell lines, and the score also ranks predictions for drug perturbations
  (\S\ref{sec:results_deployable},~\ref{sec:results_generalization}).

\end{itemize}
\endgroup

\section{Related work}
\label{sec:related}

Perturbation prediction has evolved through several paradigms~\citep{dimitrov2026interpretation}: linear models~\citep{dixit2016perturb}, VAE-based latent arithmetic~\citep{lotfollahi2019scgen,lotfollahi2023predicting}, optimal transport~\citep{bunne2023learning}, graph-based approaches with regulatory priors~\citep{roohani2024predicting}, and diffusion in expression space~\citep{he2026squidiff}; a parallel line frames perturbation response as causal inference and identifiable-representation learning~\citep{lopez2023learning,zhang2023identifiability,dong2023causal}. Most recently, single-cell foundation models (scFMs) introduce mask-prediction transformers~\citep{theodoris2023transfer,cui2024scgpt} and architectures that couple a cell-state encoder with a state-transition decoder (SE+ST factorizations)~\citep{adduri2025predicting,tahoex1_2025}, with concurrent work pursuing in-context conditioning~\citep{dong2026stack}, agentic architecture search~\citep{cheng2026harnessing}, ranking-based pretraining~\citep{taggart2026gpl}, and generative diffusion~\citep{wang2026xcell,yuan2026perturbdiff}. Yet recent benchmarks find that linear baselines match scFM-decoder performance on standard splits~\citep{ahlmann2025perturbation,csendes2025benchmarking,wenteler2024perteval,wong2025simple}, and \citet{vinastorne2026systema} show that a predictor can score well on standard metrics by capturing response patterns shared across perturbations. We open the ST decoder hidden state and trace this parity to a low-rank bottleneck inside ST; by pairing the same ST architecture with scGPT, SE, or Tahoe-x1 as the upstream encoder, we test whether the bottleneck is encoder-side or ST-side, finding the latter (\S\ref{sec:results_mechanism}).

Mechanistic interpretability provides activation patching~\citep{meng2022rome,wang2022interpretability,heimersheim2024interp}, subspace interventions~\citep{turner2023activation,burns2022discovering}, and sparse autoencoders~\citep{cunningham2023sparse,bricken2023monosemanticity,templeton2024scaling}, applied in biology to protein language models~\citep{simon2025interplm,adams2025mechanistic,parsan2025towards} and scFMs for cell-type analysis~\citep{schuster2025can}, and analyses of perturbation predictors have inspected attention maps~\citep{cui2024scgpt,adduri2025predicting}. We open the ST decoder with low-rank projections and activation patching. For deployment, predicted-NN-distance is a per-prediction confidence score, and Learn-then-Test~\citep{angelopoulos2025learn}, a method from the conformal prediction framework~\citep{angelopoulos2023conformal}, selects a threshold on it with a finite-sample guarantee. Extended related work is provided in Appendix~\ref{sec:supp_related_extended}.

\section{Methods}
\label{sec:methods}

\subsection{Preliminaries}
\label{sec:methods_prelim}

\paragraph{Model.}
We study predictors that couple an upstream encoder with a
state-transition (ST) decoder, henceforth \emph{encoder$+$ST}
(Figure~\ref{fig:family_interpolation}A). The encoder embeds each of
$S$ control cells. The ST decoder adds a perturbation embedding to
every cell embedding and passes the $S$ cells through eight attention
blocks $L_0, \dots, L_7$, and a gene decoder maps each cell to its
predicted post-perturbation expression. We pair three frozen encoders
with different pretraining, SE-600M~\citep{adduri2025predicting},
scGPT~\citep{cui2024scgpt}, and Tahoe-x1~\citep{tahoex1_2025}, each
with an ST decoder of the
same architecture and training recipe. This gives three predictors,
SE$+$ST, scGPT$+$ST, and Tahoe$+$ST. We denote by
$X_p \in \mathbb{R}^{S \times H}$ the output of the self-attention
sub-layer of $L_0$ for perturbation $p$, before it is added to the
residual stream.

\paragraph{Data and protocol.}
We use the four cell lines of the Replogle--Nadig CRISPRi
panel~\citep{replogle2022mapping,nadig2024transcriptome} (HepG2,
Jurkat, RPE1, and K562) under the STATE few-shot
protocol~\citep{adduri2025predicting}. For each target cell line, the
decoder is trained on all perturbations of the other three cell lines
and on the training perturbations of the target line, and it is tested
on the remaining perturbations of the target line. Three encoders and
four target lines give twelve encoder--cell configurations. Encoder
hidden dimensions, HVG count, context size, decoder hyperparameters,
and bootstrap settings are listed in
Table~\ref{tab:audit_config}~(\S\ref{sec:methods_setup}).

\paragraph{Notation and metric.}
For perturbation $p$ in cell line $c$, the model receives $S$ control
cells (cells with a non-targeting sgRNA) and predicts their
post-perturbation expression profiles. Averaging the $S$ predicted profiles across cells, on the highly variable gene (HVG) evaluation set $G$, gives
the pseudobulk prediction $\hat{y}_p \in \mathbb{R}^{|G|}$. Averaging
measured cells the same way gives the observed pseudobulk $y_p$ of
perturbation $p$ and the control pseudobulk $\mu^{\mathrm{ctrl}}_c$ of
cell line $c$, and $\hat{\mu}^{\mathrm{ctrl}}_c$ is the model's
prediction for the control condition. We measure accuracy by the
per-perturbation \emph{Pearson delta correlation}
(PDC)~\citep{adduri2025predicting},
\begin{equation}
\mathrm{PDC}(p, c) = \mathrm{Pearson}\!\left(\hat{y}_p - \hat{\mu}^{\mathrm{ctrl}}_c,\ y_p - \mu^{\mathrm{ctrl}}_c\right),
\label{eq:pcc_delta}
\end{equation}
the Pearson correlation across genes between the predicted and the
observed change relative to control. For comparisons between
configurations evaluated on the same perturbations, we report paired
per-perturbation differences with $95\%$ percentile-bootstrap
confidence intervals over perturbations.

\subsection{Mechanism audit}
\label{sec:methods_audit}

We characterize the ST decoder's prediction mechanism in
{\em three} increasingly specific steps. \emph{First}, a linear
or MLP probe, trained only on training perturbations, asks how much
of encoder$+$ST's per-prediction quality can be recovered from the
decoder's inputs alone. \emph{Second}, given high recovery, we locate
the prediction inside the decoder by projecting the output of its first
attention block onto a few directions, chosen per perturbation or fixed
across perturbations. \emph{Third}, we test whether the leading
direction found this way is shared across encoders by moving it between
models trained with different encoders.

\paragraph{Probing the decoder.}
For perturbation $p$ in batch $g$, we take three components of the
trained ST checkpoint, the mean encoder embedding of the batch's
control cells, the decoder's learned embedding of $p$, and its learned
embedding of $g$, and concatenate them,
\begin{equation}
x_{p,g} = \big[\,\texttt{basal\_mean\_ctrl}(g)\,;\ \texttt{pert\_enc}(p)\,;\ \texttt{batch\_enc}(g)\,\big].
\label{eq:probe_feature}
\end{equation}
On training perturbations only, we fit a closed-form ridge regression,
with $\lambda$ chosen by grouped cross-validation on PDC, and a two-layer MLP from $x_{p,g}$ to the mean
expression of the cells carrying $p$ in batch $g$. We aggregate their predictions per held-out
perturbation, score them by PDC (Eq.~\eqref{eq:pcc_delta}), and report
the recovery ratio
$\mathrm{PDC}(\text{probe})/\mathrm{PDC}(\text{enc}+\text{ST})$.
Appendix~\ref{sec:methods_probes} gives the probe settings, the
simpler baselines we compare against, the checks that test
perturbations do not leak into training, and two larger probes.

\paragraph{Localizing the prediction.}
To locate the prediction inside the decoder, we compute the SVD of the
$L_0$ attention output, $X_p = U_p \Sigma_p V_p^\top$, and truncate it
to its top-$k$ singular components,
\begin{equation}
\hat{X}_p^{(k)} = U_p^{(k)} \Sigma_p^{(k)} V_p^{(k)\top},\qquad k\in\{1,4,8,16,32\},
\label{eq:rank_k_svd}
\end{equation}
substituting $\hat{X}_p^{(k)}$ for $X_p$ at inference time and
re-scoring PDC (\S\ref{sec:methods_rank_k} describes the same
intervention at $L_1$--$L_7$). As a control, we instead keep $k$ random
orthonormal directions (\S\ref{sec:methods_rank_k}). Both projections rescale each row to its original norm, so
they differ only in which $k$-dimensional subspace is kept. If the
top-$k$ truncation preserves PDC and the random control does not, the
prediction depends on the leading singular directions.

\paragraph{Testing a subspace shared across perturbations.}
The truncation above takes its directions from each perturbation's own
activation. To test whether one subspace serves all perturbations, we
also project onto a basis estimated once from training: the top-$k$
eigenvectors $V^{\mathrm{train}}_k$ of
$M = |\mathcal{T}|^{-1}\sum_{q\in\mathcal{T}} n_q^{-1} X_q^\top X_q$,
where $\mathcal{T}$ is the set of training perturbations of the target
line and $n_q$ is the number of rows of $X_q$. Each test activation
becomes $X_p V^{\mathrm{train}}_k V^{\mathrm{train}\top}_k$, with rows
rescaled as above. We report the \emph{sufficient rank}, the smallest
$k$ at which the fixed basis preserves PDC and outperforms a random
basis of the same rank (test in \S\ref{sec:methods_sufficient_rank}).

\paragraph{Testing a direction shared across encoders.}
We move the leading direction between two models trained on the same
cell line with different encoders, a donor and a recipient. In the
recipient, we replace the $i$-th singular component
$\sigma_i\, u_i\, v_i^\top$ of $X_p$ at $L_0$ with
\begin{equation}
\sigma_i^{(\mathrm{recipient})}\, u_i^{(\mathrm{recipient})}\, v_i^{(\mathrm{donor})\top},
\label{eq:patch}
\end{equation}
keeping the recipient's $u_i$ and $\sigma_i$ and taking $v_i$ from the
donor's forward pass on the same perturbation. The recipient continues
through the remaining attention blocks, and we compare its PDC with
that of the recipient without the intervention, perturbation by perturbation. We patch
each of the top three components ($i = 1, 2, 3$) for every encoder pair
and cell line. Variants that rescale the patched component, and
ablations that separate its direction from its magnitude, are in
\S\ref{sec:methods_patching}. Patching from a donor trained on a
different cell line separates encoder-specific from cell-specific
structure (\S\ref{sec:bundle_g_wrongcell}).

\subsection{Per-prediction confidence score}
\label{sec:methods_trust_signal}

PDC declines as the cosine distance from a perturbation's observed
response to its nearest training response increases
(Figure~\ref{fig:family_interpolation}B). We call this the
\emph{oracle distance},
\begin{equation}
d_{\mathrm{NN}}(p) = \min_{q\in\mathrm{train}} \mathrm{cos\_dist}\!\left(\Delta^{\mathrm{obs}}_p,\ \Delta^{\mathrm{train}}_q\right),
\label{eq:nn_dist}
\end{equation}
where $\Delta^{\mathrm{obs}}_p = y_p - \mu^{\mathrm{ctrl}}_c$, the
training responses $\Delta^{\mathrm{train}}_q = y_q - \mu^{\mathrm{ctrl}}_c$
range over the training perturbations $q$ of cell line $c$, and
$\mathrm{cos\_dist}(a,b) = 1 - a^\top b/(\|a\|_2\|b\|_2)$.
Predicted-NN-distance computes the same distance from the model's
predicted response, so it is available before the perturbation is
measured.

\begin{definition}[Predicted-NN-distance]
\label{def:pred_nn_dist}
The \emph{predicted-NN-distance} of a test perturbation $p$ in cell
line $c$ is
\begin{equation}
d_{\mathrm{NN}}^{\mathrm{pred}}(p) = \min_{q\in\mathrm{train}} \mathrm{cos\_dist}\!\left(\Delta^{\mathrm{pred}}_p,\ \Delta^{\mathrm{train}}_q\right),
\qquad \Delta^{\mathrm{pred}}_p = \hat{y}_p - \hat{\mu}^{\mathrm{ctrl}}_c.
\label{eq:pred_nn_dist}
\end{equation}
\end{definition}

\paragraph{Selective prediction.}
Following selective prediction~\citep{geifman2017selective}, a
prediction is accepted when its score is small.

\begin{definition}[Accepted prediction]
\label{def:trust_regime}
The prediction for perturbation $p$ is \emph{accepted} if
$d_{\mathrm{NN}}^{\mathrm{pred}}(p) \leq \tau$. We use one $\tau$ for all
encoder--cell configurations, chosen to make the accepted predictions
as accurate as possible relative to the rank-$1$ SVD baseline, averaged
over configurations (Appendix~\ref{sec:supp_pooled_tau}).
\end{definition}
The rank-$1$ SVD baseline is the accuracy obtained by capturing only
the response pattern shared by all perturbations: the mean Pearson
correlation between each observed held-out response and its
reconstruction from the leading singular direction of the observed
held-out responses (Table~\ref{tab:target_rank}). In deployment, a
practitioner computes the score from the prediction and the cached
table of training responses, and acts on the prediction if the score
is at most $\tau$.

\paragraph{Evaluation.}
We evaluate the score by its Spearman correlation with per-perturbation
PDC, reported as $|\rho|$, with the same correlation for the oracle
distance as a reference. We then accept held-out predictions in order
of increasing score and report the mean PDC of the accepted predictions
as a function of the share accepted, and the mean PDC of the accepted
and rejected predictions under the threshold of
Definition~\ref{def:trust_regime}.
Appendix~\ref{sec:supp_pred_nn_magnitude_stratified} tests whether the
score carries information beyond the size of the predicted response.

\section{Results}
\label{sec:results}

In this section we compare the encoder$+$ST predictors with simple
baselines and relate their accuracy to distance from training
(\S\ref{sec:results_trust_boundary}), locate where the decoder forms its
prediction (\S\ref{sec:results_mechanism}), and evaluate
predicted-NN-distance on these predictors
(\S\ref{sec:results_deployable}) and on other predictors and
perturbation types.

\subsection{Benchmark accuracy and distance to training}
\label{sec:results_trust_boundary}
\label{sec:results_benchmark}

\begin{table}[t]
\centering
\small
\caption{\textbf{Comparison with simple predictors on held-out
perturbations.} Mean PDC (Eq.~\eqref{eq:pcc_delta}) over held-out
perturbations and over those far from every training response (cosine
distance above $0.5$). The decoders score in the range of the simple
predictors, and every predictor loses most of its accuracy far from
training. Per-cell-line far values, counts, and further baselines are in
Appendix~\ref{sec:supp_rev_comparison}.}
\label{tab:comparison_main}
\begin{tabular}{@{}ll*{5}{c}@{\hspace{10pt}}c@{}}
\toprule
 & & \multicolumn{5}{c}{held-out perturbations} & far \\
\cmidrule(lr){3-7}\cmidrule(l){8-8}
 & & HepG2 & Jurkat & RPE1 & K562 & mean & mean \\
\midrule
\multicolumn{2}{@{}l}{Train-mean response} & 0.293 & 0.171 & 0.416 & 0.148 & 0.257 & 0.073 \\
\multicolumn{2}{@{}l}{CPA~\citep{lotfollahi2023predicting}} & 0.403 & 0.330 & 0.501 & 0.302 & 0.384 & 0.117 \\
\addlinespace[4pt]
SE$+$ST & \emph{full} & 0.388 & 0.306 & 0.475 & 0.327 & 0.374 & 0.093 \\
 & \emph{rank-1} & 0.385 & 0.298 & 0.474 & 0.331 & 0.372 & 0.092 \\
 & \emph{ridge} & 0.369 & 0.334 & 0.473 & 0.331 & 0.377 & 0.148 \\
\addlinespace[4pt]
Tahoe$+$ST & \emph{full} & 0.442 & 0.376 & 0.553 & 0.382 & 0.438 & 0.140 \\
 & \emph{rank-1} & 0.430 & 0.387 & 0.535 & 0.389 & 0.435 & 0.142 \\
 & \emph{ridge} & 0.396 & 0.360 & 0.490 & 0.348 & 0.398 & 0.159 \\
\addlinespace[4pt]
scGPT$+$ST & \emph{full} & 0.364 & 0.309 & 0.477 & 0.196 & 0.337 & 0.071 \\
 & \emph{rank-1} & 0.295 & 0.117 & 0.465 & 0.182 & 0.265 & 0.039 \\
 & \emph{ridge} & 0.379 & 0.336 & 0.460 & 0.291 & 0.367 & 0.140 \\
\bottomrule
\end{tabular}
\end{table}

\paragraph{Baselines.}
We compare each encoder$+$ST predictor with three simpler predictors: a
ridge regression on the decoder's input embeddings
(Eq.~\eqref{eq:probe_feature}), CPA~\citep{lotfollahi2023predicting},
an earlier autoencoder-based predictor, and the train-mean response,
which predicts the mean training response for every perturbation.
Table~\ref{tab:comparison_main} also reports each decoder with its
$L_0$ attention output reduced to its top singular component, which
\S\ref{sec:results_mechanism} uses.

\paragraph{Benchmark accuracy.}
Table~\ref{tab:comparison_main} compares the predictors over all
held-out perturbations and over those far from every training response.
The ridge regression reads the same three embeddings that the decoder
receives, so the gap between the two is what the decoder's attention
over control cells contributes. Over all held-out perturbations this gap
is small. Each decoder and its ridge regression score within $0.04$ PDC
of each other, and CPA falls in the same range. On perturbations far
from every training response, the decoders drop to $0.07$--$0.14$ PDC,
close to the $0.07$ of the train-mean response, and each falls below its
ridge regression. Most of the accuracy these predictors show on the
benchmark therefore comes from perturbations that resemble training
perturbations.

\paragraph{Distance to training.}
Figure~\ref{fig:family_interpolation}B shows the same effect as a
continuous relationship. Per-perturbation PDC declines smoothly with the
oracle distance (Eq.~\eqref{eq:nn_dist}), and the curve has the same
shape for all three encoders and in all four cell lines
(Figure~\ref{fig:supp_trust_boundary_4cells}; Spearman $\rho$ from
$-0.92$ to $-0.73$). A perturbation's position on this curve
therefore indicates how accurate its prediction will be, whichever
encoder is used. If this distance can be estimated before the
experiment, it can indicate which predictions to accept
(\S\ref{sec:results_deployable}).
The two examples from \S\ref{sec:intro} lie at the two ends of the
curve. \gene{RPL6} has close ribosomal neighbors in training and is
predicted accurately with every encoder. \gene{SNAPC5} has no close
neighbor and is predicted poorly.

\subsection{Where the decoder forms its prediction}
\label{sec:results_mechanism}

\begin{figure}[t]
\centering
\includegraphics[width=\linewidth]{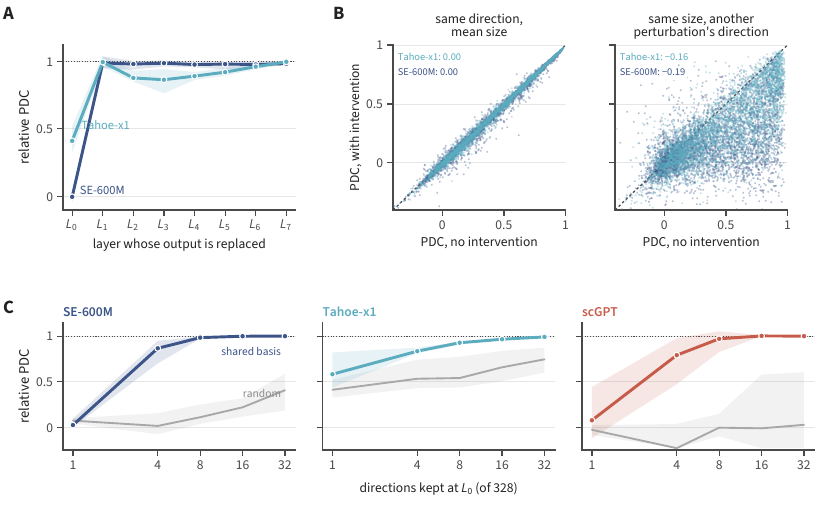}
\caption{\textbf{How the decoder reads the perturbation.} (A) PDC after
replacing the output of one attention layer with a single random
direction, relative to the decoder without the intervention (mean over
the four cell lines; band, range). (B) Per-perturbation PDC with and
without an intervention on the top $L_0$ component, which keeps its
direction and sets its size to the mean over perturbations (left), or
keeps its size and takes the direction of another perturbation (right).
Numbers give the mean change. (C) PDC after projecting the $L_0$ output
onto $k$ directions shared by all perturbations and estimated from
training activations, or onto $k$ random directions, relative to the
decoder without the intervention. For scGPT$+$ST one direction is too
few even for each perturbation's own top direction
(Table~\ref{tab:comparison_main}). The perturbation enters at $L_0$ as a
direction in a small shared subspace, and the perturbation-specific
signal lies in this direction.}
\label{fig:hero}
\end{figure}

We next locate where in the decoder this dependence on the training
perturbations arises. We intervene on the attention outputs of a trained decoder (\S\ref{sec:methods_audit}) and measure the change in PDC.

\paragraph{Where the perturbation enters.}
For one perturbation, the output of the decoder's first attention layer
over the $S$ control cells is close to rank one
(Appendix~\ref{sec:rank_collapse}),
$X_p \approx \sigma_1 u_1 v_1^\top$: the layer adds nearly the same
vector $v_1$ to every control cell, scaled by a cell-specific weight
$u_{1i}$ (Figure~\ref{fig:family_interpolation}A; Appendix~\ref{sec:supp_biology}
describes the genes along this direction). Keeping only this
component changes mean PDC by at most $0.003$ for SE$+$ST and Tahoe$+$ST
(the rank-1 rows of Table~\ref{tab:comparison_main}). Replacing the
layer's output with a random direction removes the PDC of SE$+$ST and
more than half of the PDC of Tahoe$+$ST, while the same replacement at
any of $L_1$ to $L_7$ leaves PDC close to that of the decoder without the intervention
(Figure~\ref{fig:hero}A). The perturbation-specific signal therefore
enters at the first attention layer. Setting its output to zero brings
SE$+$ST to a PDC near zero and costs Tahoe$+$ST about a third of its PDC,
so part of the signal of Tahoe$+$ST reaches the output by another route
(Appendix~\ref{sec:what_is_ST_doing}).

\paragraph{The direction carries the signal.}
Setting the size of the top component to its mean over perturbations
leaves PDC unchanged for SE$+$ST and Tahoe$+$ST, while giving each
perturbation the direction of another perturbation lowers it
(Figure~\ref{fig:hero}B). The prediction then follows the direction it
received. For pairs whose two perturbations are both predicted well and
have dissimilar responses, the prediction after the swap is closer to
the observed response of the perturbation that supplied the direction
than to its own in $90\%$ of Tahoe$+$ST pairs and in all SE$+$ST pairs
(Appendix~\ref{sec:t11_orientation}). The direction is also specific to
the encoder and the cell line. Substituting another encoder's direction
lowers PDC by $0.1$ to $0.3$ in most cell lines
(Appendix~\ref{sec:cross_encoder_patching}), and substituting the
direction of the same encoder trained on another cell line lowers it by
$0.16$ to $0.23$ for Tahoe$+$ST in every cell line, and by less for
SE$+$ST and scGPT$+$ST (Appendix~\ref{sec:bundle_g_wrongcell}).

\paragraph{A small shared subspace.}
Across perturbations, these directions lie in a small subspace that can
be estimated from training data. A single basis estimated from training
activations and shared by all perturbations preserves PDC with at most
$32$ of the $328$ dimensions, while the same number of random directions
does not (Figure~\ref{fig:hero}C). The shared basis needs more
directions than each perturbation's own top directions, because
different perturbations use different directions within it. scGPT$+$ST
needs a few components even for its own top directions.

\paragraph{Retrieval from the training responses.}
To measure what the training data could support, we use an oracle that
never runs the decoder. For a held-out perturbation, it finds the five
training perturbations whose observed responses are closest to the
perturbation's own measured response and predicts their weighted
average. This oracle scores above the decoder in every configuration, by
$0.137$ PDC on average (Appendix~\ref{sec:nn_oracle}). It needs the
measured response and cannot be used before an experiment, but it shows
that the decoder recovers less than the nearest training responses
already contain. Because the first attention layer passes the perturbation through directions fixed by the training perturbations, a held-out
perturbation is predicted well when its response resembles a training
response, which is the dependence on distance in
\S\ref{sec:results_trust_boundary}.

\subsection{A per-prediction confidence score}
\label{sec:results_deployable}

\begin{figure}[t]
\centering
\includegraphics[width=\linewidth]{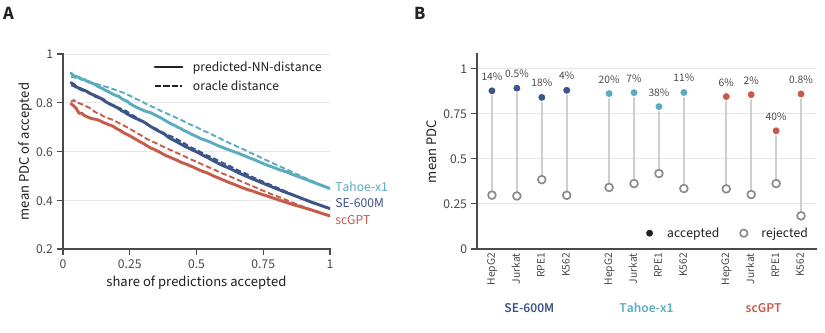}
\caption{\textbf{Selecting predictions with predicted-NN-distance.} (A)
Mean PDC of the accepted predictions when held-out predictions are
accepted in order of increasing distance, averaged over the four cell
lines of each encoder; solid, predicted-NN-distance; dashed, the oracle
distance from the observed response. At share $1$ every curve reaches
the mean PDC of all predictions. (B) Mean PDC of the accepted and
rejected predictions under the pooled threshold $\tau \approx 0.10$ in
each encoder--cell configuration; numbers give the share accepted. The
score selects predictions nearly as well as the oracle, and one threshold
selects accurate predictions in every configuration.}
\label{fig:universal_threshold}
\end{figure}

The distance in \S\ref{sec:results_benchmark} is computed from the
observed response, so it is unavailable before an experiment. We test
whether predicted-NN-distance (\S\ref{sec:methods_trust_signal}), computed from the prediction and the training responses, can take its place when deciding which
predictions to accept.

\paragraph{Ranking predictions.}
Predicted-NN-distance ranks predictions almost as well as the oracle
distance (Spearman $|\rho| = 0.81$ against $0.88$, averaged over
configurations; Appendix~\ref{sec:supp_predicted_nn_followups}).
Figure~\ref{fig:universal_threshold}A shows the mean PDC of the accepted
predictions as more of them are accepted in increasing order of
distance. For every encoder the curve for predicted-NN-distance nearly
coincides with that of the oracle distance, and the tenth of predictions
with the smallest distance is about twice as accurate as the average
prediction. The score can therefore replace the distance of
\S\ref{sec:results_benchmark} when the observed response is not
available.

\paragraph{Beyond the size of the prediction.}
Predictions with a large predicted response tend to be more accurate. Among predictions of similar size the score still ranks them by PDC, and it ranks them better than the distance to
the ten nearest training responses, the Euclidean distance, and the
alignment with the leading axis of the training responses
(Appendices~\ref{sec:supp_score_baselines}
and~\ref{sec:supp_pred_nn_magnitude_stratified}).

\paragraph{One threshold for all configurations.}
A threshold fitted for each configuration needs observed responses from
that configuration. We therefore use one threshold, $\tau \approx 0.10$,
for all configurations (selection rule in
Appendix~\ref{sec:supp_pooled_tau}). In every configuration the accepted
predictions are far more accurate than the rejected ones
(Figure~\ref{fig:universal_threshold}B). The share accepted is smallest
in Jurkat and K562, which have the fewest held-out perturbations with a
close training response. A threshold refitted without one cell line or
one encoder gives nearly the same accuracy on the left-out one
(Appendix~\ref{sec:supp_pooled_tau}), so a threshold fitted on the other
configurations can be used on a left-out cell line or encoder. Under this
threshold \gene{RPL6} is accepted and \gene{SNAPC5} is rejected with
every encoder.

\paragraph{A threshold with a guarantee.}
When a target accuracy is set in advance,
Learn-then-Test~\citep{angelopoulos2025learn} chooses a threshold whose
accepted predictions meet the target with a finite-sample guarantee,
provided the calibration and deployment perturbations come from the same
population. In our setting it certifies all but one configuration
(Appendix~\ref{sec:supp_rev_ltt}). Like a threshold fitted for each
configuration, it needs observed responses from that configuration.

\subsection{Beyond the audited setting}
\label{sec:results_generalization}

\begin{figure}[t]
\centering
\includegraphics[width=\linewidth]{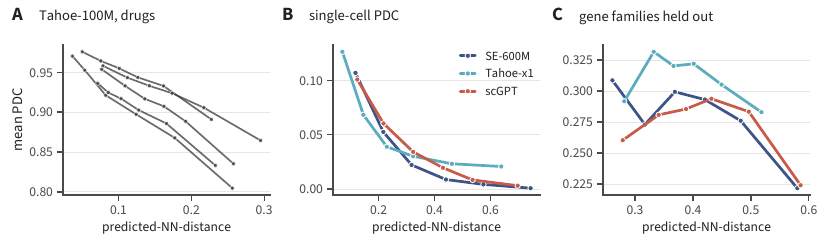}
\caption{\textbf{Predicted-NN-distance outside the audited setting.}
Mean PDC in six equal-count bins of predicted-NN-distance. (A) Arc's ST
model on Tahoe-100M, drug--dose conditions in five test cell lines, one
line per cell line. (B) PDC computed against single cells instead of the
pseudobulk, one line per encoder, mean over cell lines. (C) ST decoders
retrained with one gene family held out, scored on that family, one line
per encoder, mean over cell lines. The score correlates with PDC in A and B
(Spearman $|\rho| = 0.57$--$0.85$) and more weakly after family-held-out
retraining (C; negative in $9$ of $12$ configurations).}
\label{fig:generalization}
\end{figure}

We now test the account of \S\ref{sec:results_benchmark}--\ref{sec:results_deployable}
when the perturbation type, the resolution of the evaluation, and the
split change (Figure~\ref{fig:generalization}).

\paragraph{Drug perturbations.}
On Tahoe-100M~\citep{zhang2025tahoe} we ran the ST model that Arc
Institute released for this dataset, without retraining, on the
drug--dose conditions of its five test cell lines. The score correlates with PDC
in all five (Figure~\ref{fig:generalization}A), also when references to
the same drug are removed from the training pool
(Appendix~\ref{sec:supp_rev_tahoe}). The model has seen most of these
drugs in other cell lines, so this tests distance to training within a
cell line rather than drugs the model has never seen.

\paragraph{Single-cell PDC.}
PDC computed against individual cells instead of the pseudobulk also
falls with predicted-NN-distance for all three encoders
(Figure~\ref{fig:generalization}B). For Tahoe$+$ST part of this
relationship reflects the size of the response, since dividing each
perturbation's single-cell PDC by the highest value its cells allow
weakens the correlation (Appendix~\ref{sec:supp_rev_singlecell}).

\paragraph{Gene families held out.}
To test whether accuracy depends on distance or on the presence of the
same gene family in training, we removed each of eight gene families
from training and validation in all four cell lines, retrained the
decoder, and scored only the held-out family. The score stays
negatively correlated with PDC in nine of the twelve configurations
(Figure~\ref{fig:generalization}C). In the three RPE1 configurations the
correlation is positive, and its interval includes zero
(Appendix~\ref{sec:supp_rev_family_holdout}). When the nearest training
perturbations come from other families, distance ranks predictions less
reliably than on random held-out perturbations.

\section{Discussion}
\label{sec:discussion}

In this work we found that the encoder$+$ST predictors are reliable for
a perturbation when the perturbation's response is close to a response
seen in training. We traced this dependence to the decoder's first
attention layer, where each perturbation is written as one direction in
a small subspace estimated from training perturbations. These results
suggest why perturbation models built on foundation-model
encoders often do no better than linear or mean
baselines~\citep{ahlmann2025perturbation,csendes2025benchmarking,wong2025simple,vinastorne2026systema}.
Projecting the first-layer output onto a few directions leaves PDC unchanged, and ridge regression on the same embeddings scores about the same as the full decoder (Table~\ref{tab:comparison_main}, Figure~\ref{fig:hero}C). Both predict well when a
held-out response resembles a training response and poorly otherwise,
so a benchmark score depends largely on how many test perturbations lie
close to training.

Replacing a perturbation's first-layer direction with another
perturbation's direction makes the decoder predict the second
perturbation's response (Appendix~\ref{sec:t11_orientation}). Similar interventions
have shown that single directions mediate behaviors in language
models~\citep{arditi2024refusal}, and causal abstraction gives a general
framework for such tests~\citep{geiger2025causal}. Running them on other perturbation
decoders is a cheap first test of whether they share this structure. For users of the current models,
predicted-NN-distance needs only the prediction and the training
responses, so it can be computed before an experiment. A new cell line
or encoder rarely comes with the measurements needed to calibrate a
threshold. In our data, a threshold chosen without a cell line or an
encoder worked on each of the four cell lines and three encoders we held
out (Appendix~\ref{sec:supp_pooled_tau}). When a guarantee is needed,
Learn-then-Test provides one from measurements in the same setting.

\paragraph{Limitations and future work.}
We audited one decoder architecture with three encoders, on CRISPRi
screens in four cell lines and on drug perturbations in Tahoe-100M. The
bottleneck is clearest for SE$+$ST; in Tahoe$+$ST part of the
perturbation signal bypasses the first attention layer
(Figure~\ref{fig:hero}A; Appendix~\ref{sec:what_is_ST_doing}). The score
is weaker evidence of accuracy when a whole gene family is absent from
training, because the nearest training responses then come from other
families (\S\ref{sec:results_generalization}), and the Learn-then-Test
guarantee fails when deployment perturbations lie farther from training
than the calibration perturbations (Appendix~\ref{sec:supp_rev_ltt}). We
intervened on attention outputs but not on the MLPs of later layers, and
we have not established what each perturbation's direction encodes in
biological terms. Predicted-NN-distance also correlates with PDC for
TxPert, a GNN-based predictor (Appendix~\ref{sec:supp_rev_txpert}), and
auditing such predictors would show whether their dependence on
distance has the same cause. Predicted-NN-distance could also guide which perturbations
to test next. Perturbations with a large distance are those whose
predictions the training data support least, so they are natural
candidates for the next experiment. How to choose such experiments is
studied in active learning and experimental design for genetic
perturbations~\citep{mehrjou2022genedisco,tigas2022interventions,annadani2024amortized,zhang2026gocbed}.

\paragraph{Broader impacts.}
\label{sec:broader_impacts}
Perturbation predictors are used to choose which experiments to run.
Predicted-NN-distance is computed from the prediction and the training
data, so researchers can tell before an experiment which predictions are
likely to hold and spend experiments accordingly. A risk is that
accepted predictions are treated as measured responses, although some
of them are wrong.

\begin{ack}
All authors acknowledge support from NIH awards U24DK138515 and
OT2OD038003. The authors declare no competing interests.
\end{ack}

\newpage
\bibliographystyle{plainnat}
\bibliography{reference}

\newpage
\section*{NeurIPS Paper Checklist}

\begin{enumerate}

\item {\bf Claims}
    \item[] Question: Do the main claims made in the abstract and introduction accurately reflect the paper's contributions and scope?
    \item[] Answer: \answerYes{}
    \item[] Justification: The abstract and introduction present our key contributions: (i) a rank-1, encoder-conditioned bottleneck at the first attention layer of the state-transition (ST) decoder, and (ii) predicted-NN-distance as a deployable per-prediction trust signal, and these claims are validated through empirical studies (\S~\ref{sec:results}).

\item {\bf Limitations}
    \item[] Question: Does the paper discuss the limitations of the work performed by the authors?
    \item[] Answer: \answerYes{}
    \item[] Justification: The limitation is discussed in \S\ref{sec:discussion}.

\item {\bf Theory assumptions and proofs}
    \item[] Question: For each theoretical result, does the paper provide the full set of assumptions and a complete (and correct) proof?
    \item[] Answer: \answerYes{}
    \item[] Justification: For the definitions mentioned in the manuscript, we stated the assumptions clearly.

    \item {\bf Experimental result reproducibility}
    \item[] Question: Does the paper fully disclose all the information needed to reproduce the main experimental results of the paper to the extent that it affects the main claims and/or conclusions of the paper (regardless of whether the code and data are provided or not)?
    \item[] Answer: \answerYes{}
    \item[] Justification: Please refer to \S\ref{sec:methods} and the extended methods (\S\ref{sec:supp_methods_extended}) in Appendix for detailed information.

\item {\bf Open access to data and code}
    \item[] Question: Does the paper provide open access to the data and code, with sufficient instructions to faithfully reproduce the main experimental results, as described in supplemental material?
    \item[] Answer: \answerYes{}{}
    \item[] Justification: The data~\citep{replogle2022mapping, nadig2024transcriptome} are open-sourced, and the code is available at \url{https://github.com/HowIII/lowrank-decoder-reliability}.

\item {\bf Experimental setting/details}
    \item[] Question: Does the paper specify all the training and test details (e.g., data splits, hyperparameters, how they were chosen, type of optimizer) necessary to understand the results?
    \item[] Answer: \answerYes{}
    \item[] Justification: Please refer to \S\ref{sec:methods_setup}, \S\ref{sec:methods_probes}, and\S\ref{sec:methods_stats}.

\item {\bf Experiment statistical significance}
    \item[] Question: Does the paper report error bars suitably and correctly defined or other appropriate information about the statistical significance of the experiments?
    \item[] Answer: \answerYes{}
    \item[] Justification: Please refer to \S\ref{sec:methods_stats}. 

\item {\bf Experiments compute resources}
    \item[] Question: For each experiment, does the paper provide sufficient information on the computer resources (type of compute workers, memory, time of execution) needed to reproduce the experiments?
    \item[] Answer: \answerYes{}
    \item[] Justification: Please refer to \S\ref{sec:supp_compute}.

\item {\bf Code of ethics}
    \item[] Question: Does the research conducted in the paper conform, in every respect, with the NeurIPS Code of Ethics \url{https://neurips.cc/public/EthicsGuidelines}?
    \item[] Answer: \answerYes{}
    \item[] Justification: We have reviewed the NeurIPS Code of Ethics and confirm that our research fully complies with these guidelines.
    
\item {\bf Broader impacts}
    \item[] Question: Does the paper discuss both potential positive societal impacts and negative societal impacts of the work performed?
    \item[] Answer: \answerYes{}
    \item[] Justification: Yes. Please refer to \S\ref{sec:broader_impacts}.

\item {\bf Safeguards}
    \item[] Question: Does the paper describe safeguards that have been put in place for responsible release of data or models that have a high risk for misuse (e.g., pre-trained language models, image generators, or scraped datasets)?
    \item[] Answer: \answerNA{}
    \item[] Justification: Our work mainly focuses on understanding the mechanisms of a biological foundation model and quantifying its prediction's reliability, which does not pose risks of misuse.

\item {\bf Licenses for existing assets}
    \item[] Question: Are the creators or original owners of assets (e.g., code, data, models), used in the paper, properly credited and are the license and terms of use explicitly mentioned and properly respected?
    \item[] Answer: \answerYes{}
    \item[] Justification: 
    We properly cite all assets used in our research with their license, baseline methods with their original papers, and all open-source libraries used according to their respective licenses. All citations and attributions are included in our references section.
    
\item {\bf New assets}
    \item[] Question: Are new assets introduced in the paper well documented and is the documentation provided alongside the assets?
    \item[] Answer: \answerNA{}
    \item[] Justification: The submission introduces no new released datasets, no new released models, and no new released checkpoints. The analysis code is released with documentation at \url{https://github.com/HowIII/lowrank-decoder-reliability}.

\item {\bf Crowdsourcing and research with human subjects}
    \item[] Question: For crowdsourcing experiments and research with human subjects, does the paper include the full text of instructions given to participants and screenshots, if applicable, as well as details about compensation (if any)?
    \item[] Answer: \answerNA{}
    \item[] Justification: Our research does not involve any crowdsourcing or human subjects.

\item {\bf Institutional review board (IRB) approvals or equivalent for research with human subjects}
    \item[] Question: Does the paper describe potential risks incurred by study participants, whether such risks were disclosed to the subjects, and whether Institutional Review Board (IRB) approvals (or an equivalent approval/review based on the requirements of your country or institution) were obtained?
    \item[] Answer: \answerNA{}
    \item[] Justification: The paper does not involve human subjects. 
    
\item {\bf Declaration of LLM usage}
    \item[] Question: Does the paper describe the usage of LLMs if it is an important, original, or non-standard component of the core methods in this research? Note that if the LLM is used only for writing, editing, or formatting purposes and does \emph{not} impact the core methodology, scientific rigor, or originality of the research, declaration is not required.
    \item[] Answer: \answerYes{}
    \item[] Justification: We used LLM-based assistants for coding, analysis checks, literature search, and writing; LLMs are not part of the methods studied. See Appendix~\ref{sec:llm_usage}.

\end{enumerate}

\newpage
\appendix
\section*{\appendixname}
\addcontentsline{toc}{section}{\appendixname}
\startcontents[sections]
\begingroup\setlength{\parskip}{2pt}
\printcontents[sections]{}{1}{\setcounter{tocdepth}{2}}
\endgroup
\newpage

\section{Extended methods}
\label{sec:supp_methods_extended}

\subsection{Datasets, model variants, and evaluation protocol}
\label{sec:methods_setup}

\begin{table}[h]
\centering
\caption{Audit configurations and hyperparameters. The encoder changes
across configurations. The ST decoder architecture and training recipe
are the same in every run.}
\label{tab:audit_config}
\small
\begin{tabular}{@{}ll@{}}
\toprule
\textbf{Setting} & \textbf{Value} \\
\midrule
\multicolumn{2}{@{}l}{\emph{Encoders (frozen)}} \\
SE-600M output dimension                  & $2{,}058$ \\
scGPT output dimension                    & $512$ \\
Tahoe-x1 (\texttt{tx-70m}) output dimension & $512$ \\
\midrule
\multicolumn{2}{@{}l}{\emph{Inference}} \\
Control cells per perturbation ($S$)      & $64$ \\
Genes scored (HVGs)                       & $2{,}000$ \\
\midrule
\multicolumn{2}{@{}l}{\emph{ST decoder}} \\
Hidden dimension ($H$)                    & $328$ \\
Attention blocks                          & $8$ ($L_0,\dots,L_7$) \\
Attention heads / head dimension          & $12$ / $64$ \\
MLP intermediate size                     & $3{,}072$ \\
\midrule
\multicolumn{2}{@{}l}{\emph{Statistics}} \\
Bootstrap resamples / interval            & $5{,}000$ / $95\%$ \\
\bottomrule
\end{tabular}
\end{table}

\paragraph{Cell lines and split.}
We use the four Replogle--Nadig cell lines HepG2, Jurkat, RPE1, and
K562~\citep{replogle2022mapping,nadig2024transcriptome} with the
few-shot train, validation, and test perturbation lists of
\citet{adduri2025predicting}. The test sets have $945$ (HepG2), $1{,}084$
(Jurkat), $1{,}061$ (RPE1), and $968$ (K562) perturbations. Test
perturbations are never used to train probes or to estimate the fixed
basis. The replacement experiments of \S\ref{sec:methods_patching} act
on test perturbations and therefore use their own $L_0$ directions.
PDC is computed on the $2{,}000$ highly variable genes of
the published selection, as in \citet{adduri2025predicting}.

\paragraph{Models and inference.}
Each predictor is a frozen encoder followed by an ST decoder. The
encoder computes an embedding for every cell once, and the decoder is
trained on these cached embeddings with a one-hot perturbation
encoding, so the encoder receives no gradient from the perturbation
task. The decoder is a Llama-style transformer with bidirectional
attention and a batch encoder (Table~\ref{tab:audit_config}). SE$+$ST
uses the published few-shot checkpoints of \citet{adduri2025predicting}.
For scGPT$+$ST and Tahoe$+$ST we trained the decoder with the same
configuration, changing only the input embedding, and trained
Tahoe$+$ST on HepG2 with two further seeds.
Unless stated otherwise, the decoder receives $64$ control cells from
the same batch (gem group) as context for each perturbation and returns $64$
predicted cells, which we average into a pseudobulk prediction.
Comparisons between two predictors are paired on the test
perturbations they share.

\subsection{Hidden-state diagnostics}
\label{sec:methods_diagnostics}

For the comparison in Appendix~\ref{sec:rank_collapse}, we collect, for each perturbation $p$, the $L_0$ attention output over
its context cells, $X_p \in \mathbb{R}^{S \times H}$, and compute its
singular values $\sigma_j(X_p)$. We report the rank-1 fraction
$r_1(X_p) = \sigma_1^2 / \sum_j \sigma_j^2$ and the participation ratio
$\mathrm{PR}(X_p) = (\sum_j \sigma_j^2)^2 / \sum_j \sigma_j^4$, and
compare the trained decoder with three random initializations.

\subsection{Linear and MLP probes}
\label{sec:methods_probes}

The probes measure how much of the decoder's accuracy a small model
recovers from the same conditioning inputs. The ridge probe is the ridge
baseline of Table~\ref{tab:comparison_main}, and the other probes are
reported in Appendix~\ref{sec:b_prime}. For test perturbation $p$
in gem group $g$, the input is the mean control-cell embedding of $g$,
the perturbation encoding of $p$, and the batch encoding of $g$, with
the two encodings taken from the trained decoder (Eq.~\eqref{eq:probe_feature}).
The probe sees the mean of the control cells, while the decoder sees
the $64$ cells themselves.

Training rows are pairs of a training perturbation and a gem group with
at least three cells, and test rows are pairs of a test perturbation
and a gem group, so no test perturbation contributes to training. The
main probe is a two-layer MLP. We also fit ridge regression on the
same input, ridge on the control and batch inputs only, ridge on the
perturbation and batch inputs only, ridge with shuffled perturbation
labels, and a constant prediction equal to the control mean. Two
further probes test whether a larger model or an explicit interaction
between control state and perturbation changes the result: an MLP with
about as many parameters as the decoder's MLP layers, and a probe that
adds a low-rank product of the control and perturbation inputs to a
linear model. Table~\ref{tab:probe_hparams} lists the settings. Probe
predictions are scored with PDC on the same $2{,}000$ genes as the
decoder.

\begin{table}[h]
\centering
\caption{Probe settings.}
\label{tab:probe_hparams}
\small
\begin{tabular}{@{}lp{0.62\linewidth}@{}}
\toprule
Probe & Settings \\
\midrule
MLP (main) & 2 layers, hidden $256$, ReLU, dropout $0.1$, Adam, learning rate $10^{-3}$, batch $256$, $50$ epochs \\
Ridge & closed form, penalty chosen by $5$-fold grouped cross-validation on PDC \\
Large MLP & 3 layers, hidden $2{,}048$ ($17$--$23$M parameters), GELU, dropout $0.1$, AdamW with cosine schedule, $120$ epochs, early stopping on $10\%$ of training rows (patience $15$) \\
Interaction & linear term plus $B^\top(\phi_b(\text{control}) \odot \phi_p(\text{perturbation}))$, rank $r \in \{0, 4, 16, 64\}$, $\phi$: 2-layer MLPs (hidden $128$, GELU, dropout $0.1$), AdamW (learning rate $5{\times}10^{-4}$, weight decay $10^{-5}$), $120$ epochs, early stopping as above \\
\bottomrule
\end{tabular}
\end{table}

\subsection{Projection and ablation of the \texorpdfstring{$L_0$}{L0} output}
\label{sec:methods_rank_k}

\paragraph{Keeping the top directions.}
During the forward pass we replace each perturbation's $L_0$ attention
output $X_p$ with its truncation to the top $k$ singular directions
(Eq.~\eqref{eq:rank_k_svd}), for $k \in \{1, 4, 8, 16, 32\}$, rescale
each row to its original norm, and score PDC (the rank-1 rows of
Table~\ref{tab:comparison_main}). The random control keeps
$k$ random orthonormal directions instead, with the same rescaling,
drawn either from the whole space (SE$+$ST and scGPT$+$ST in
Figure~\ref{fig:hero}C) or from the subspace orthogonal to the top $32$
singular directions of $X_p$ (Tahoe$+$ST and the layer scan). The layer scan applies
the same top-1 truncation and the same random control to the attention
output of one layer at a time, $L_0$ to $L_7$
(Figure~\ref{fig:hero}A, Appendix~\ref{sec:layer_scan}). The ablation
sets the $L_0$ attention output to zero
(Appendix~\ref{sec:layer_scan}).

\paragraph{Fixed training basis and sufficient rank.}
\label{sec:methods_sufficient_rank}
The fixed basis is estimated from the $L_0$ outputs of the target
line's training perturbations only (\S\ref{sec:methods_audit}). For each
$k$ we compare its PDC with that of the unprojected decoder,
perturbation by perturbation, and compute a paired bootstrap interval.
The fixed basis preserves PDC at rank $k$ if this interval lies within
$\pm 0.02$ (the margin of Appendix~\ref{sec:methods_margin}), and it outperforms a random orthonormal basis of the same
rank if the interval of their difference lies above $0.02$. The
sufficient rank is the smallest $k$ at which the fixed basis meets both
conditions and the per-perturbation truncation also preserves PDC.

\subsection{Replacing directions at \texorpdfstring{$L_0$}{L0}}
\label{sec:methods_patching}

\paragraph{Directions from another encoder.}
For two predictors trained on the same cell line, we run the recipient
and replace the component $\sigma_i u_i v_i^\top$ of $X_p$ with one
that keeps the recipient's $\sigma_i$ and $u_i$ and takes $v_i$ from the
donor model's $L_0$ output for the same perturbation
(Eq.~\eqref{eq:patch}). We replace $i \in \{1, 2, 3\}$ or the components
$2$ and $3$ together (Appendix~\ref{sec:cross_encoder_patching}).

\paragraph{Size and direction of the top component.}
For Tahoe$+$ST and SE$+$ST we change one property of the top component
$\sigma_1 u_1 v_1^\top$ at a time. We set $\sigma_1$ to its mean over
perturbations and keep the direction, or keep $\sigma_1$ and replace
$v_1$ by the direction of another perturbation, assigned by a fixed
random permutation without fixed points. For Tahoe$+$ST we also replace
$v_1$ by the SE-600M direction of the same perturbation. All runs of a
configuration use the same checkpoint and the same sampled control
cells, and are compared with a run without the intervention under the
same conditions. For the direction swap we also compare the prediction for
perturbation $p$ with the observed response and the prediction without the intervention
of the perturbation whose direction it received (Figure~\ref{fig:hero}B,
Appendix~\ref{sec:t11_orientation}).

\paragraph{Directions from another cell line.}
We replace the top direction with the direction of the same encoder and
perturbation in a decoder trained on another cell line, for all three
encoders, four recipient cell lines, and the three other cell lines as
donors ($36$ combinations;
Appendix~\ref{sec:bundle_g_wrongcell}). As a check of the procedure, we also run
the replacement with the recipient's own cell line as donor ($12$
combinations), which should leave PDC unchanged.

\subsection{Confidence score and thresholds}
\label{sec:methods_thresholds}

The pooled threshold of \S\ref{sec:results_deployable} is chosen as
described in Appendix~\ref{sec:supp_pooled_tau}, and the Learn-then-Test
threshold as described in Appendix~\ref{sec:supp_rev_ltt}. The per-configuration threshold used in
Appendix~\ref{sec:supp_rev_transfer} is fitted by isotonic regression of
held-out PDC on predicted-NN-distance, as the smallest distance at which
the fitted PDC falls to the rank-1 SVD baseline
(\S\ref{sec:methods_trust_signal}) plus $0.02$.

\subsection{Statistical analysis}
\label{sec:methods_stats}

A comparison between two conditions is the mean per-perturbation
difference in PDC on the shared perturbations, with a paired $95\%$
percentile bootstrap interval over perturbations ($5{,}000$
resamples). As a check we also resample whole gene families (the nine
groups of Appendix~\ref{sec:supp_biology}), which roughly doubles the
interval widths (Appendices~\ref{sec:supp_rev_comparison}
and~\ref{sec:b_prime}).
\label{sec:methods_margin}Rerunning the decoder without any intervention, with different sampled
control cells or with a checkpoint from a different training step,
changes mean PDC by up to about $0.02$. We therefore do not interpret
differences of this size, and use $0.02$ PDC as the margin in the
sufficient rank (Appendix~\ref{sec:methods_sufficient_rank}) and in the
per-configuration threshold (Appendix~\ref{sec:methods_thresholds}).

\subsection{Compute resources}
\label{sec:supp_compute}

Experiments ran on NVIDIA L40S (48~GB) and H100 (80~GB) GPUs. Training
one ST decoder takes $4$ to $8$ hours on one GPU, and the
family-held-out experiment of Appendix~\ref{sec:supp_rev_family_holdout}
trains $108$ of them. One run of the $L_0$ interventions on a
configuration takes $1.5$ to $4$ hours, and the probes about $3.5$
hours. Distances, thresholds, and statistics run on CPUs. Batch jobs for
this paper used about $780$ GPU-hours in total, including runs not
reported here. Scripts are in the code release.

\section{Benchmark accuracy and distance to training}
\label{sec:supp_benchmark}

Section~\ref{sec:results_benchmark} reports means over the four cell
lines. Here we give the values for each cell line: the comparison with
baselines on all held-out perturbations and on the far slice
(Appendix~\ref{sec:supp_rev_comparison}), and the relationship between
accuracy and distance to training
(Appendices~\ref{sec:supp_trust_boundary_4cells}
and~\ref{sec:held_out_generalization}).

\subsection{Comparison with baselines, all and far perturbations}
\label{sec:supp_rev_comparison}

Table~\ref{tab:supp_comparison_full} extends
Table~\ref{tab:comparison_main} with per-cell-line values and additional
reference rows. Every row is scored on the same held-out perturbations,
those that every predictor scores; the first row of the table gives
their number. The far slice keeps the perturbations whose observed
response is farther than $0.5$ cosine from the observed response of
every training perturbation in the $2{,}000$-HVG space. The shuffled
lookup predicts each perturbation with the response of a random training
perturbation, and the co-expression baseline weights training responses
by the control-cell correlation between their target genes and the
held-out target. Experimental reproducibility
correlates two half-sample estimates of the same response and is
therefore a reference for data noise; models are scored against the
full-sample response. On the far slice, every predictor scores at most
$0.19$ PDC in every cell line.

\begin{table}[h]
\centering\small
\setlength{\tabcolsep}{3.5pt}
\caption{Mean PDC on all held-out perturbations and on the far slice.
Same gene panel and perturbation support for every row; scVI was
scored against a separately built observed response. GEARS was scored
on K562 only.}
\label{tab:supp_comparison_full}
\resizebox{\linewidth}{!}{%
\begin{tabular}{lcccccccccc}
\toprule
 & \multicolumn{5}{c}{All held-out} & \multicolumn{5}{c}{Far ($> 0.5$ cosine)} \\
\cmidrule(lr){2-6}\cmidrule(lr){7-11}
 & K562 & RPE1 & HepG2 & Jurkat & Mean & K562 & RPE1 & HepG2 & Jurkat & Mean \\
\midrule
Perturbations ($n$) & 963 & 1{,}029 & 903 & 1{,}077 & --- & 521 & 274 & 403 & 549 & --- \\
\midrule
Shuffled-perturbation lookup & 0.072 & 0.240 & 0.151 & 0.090 & 0.138 & 0.032 & 0.034 & 0.040 & 0.044 & 0.037 \\
Train-mean response          & 0.148 & 0.416 & 0.293 & 0.171 & 0.257 & 0.065 & 0.072 & 0.078 & 0.078 & 0.073 \\
Co-expression (gene--gene)   & 0.153 & 0.419 & 0.292 & 0.169 & 0.258 & 0.067 & 0.073 & 0.077 & 0.076 & 0.073 \\
CPA~\citep{lotfollahi2023predicting} & 0.302 & 0.501 & 0.403 & 0.330 & 0.384 & 0.117 & 0.086 & 0.117 & 0.147 & 0.117 \\
scVI~\citep{lopez2018deep}  & 0.251 & 0.421 & 0.338 & 0.280 & 0.323 & 0.090 & 0.084 & 0.089 & 0.111 & 0.094 \\
GEARS~\citep{roohani2024predicting} & 0.151 & --- & --- & --- & --- & 0.045 & --- & --- & --- & --- \\
Experimental reproducibility & 0.399 & 0.617 & 0.455 & 0.411 & 0.470 & --- & --- & --- & --- & --- \\
\midrule
SE$+$ST full    & 0.327 & 0.475 & 0.388 & 0.306 & 0.374 & 0.108 & 0.082 & 0.076 & 0.106 & 0.093 \\
SE$+$ST rank-1  & 0.331 & 0.474 & 0.385 & 0.298 & 0.372 & 0.112 & 0.083 & 0.073 & 0.099 & 0.092 \\
SE$+$ST ridge   & 0.331 & 0.473 & 0.369 & 0.334 & 0.377 & 0.151 & 0.116 & 0.144 & 0.180 & 0.148 \\
Tahoe$+$ST full   & 0.382 & 0.553 & 0.442 & 0.376 & 0.438 & 0.152 & 0.119 & 0.132 & 0.156 & 0.140 \\
Tahoe$+$ST rank-1 & 0.389 & 0.535 & 0.430 & 0.387 & 0.435 & 0.159 & 0.128 & 0.119 & 0.163 & 0.142 \\
Tahoe$+$ST ridge  & 0.348 & 0.490 & 0.396 & 0.360 & 0.398 & 0.165 & 0.130 & 0.152 & 0.190 & 0.159 \\
scGPT$+$ST full   & 0.196 & 0.477 & 0.364 & 0.309 & 0.337 & 0.050 & 0.058 & 0.072 & 0.104 & 0.071 \\
scGPT$+$ST rank-1 & 0.182 & 0.465 & 0.295 & 0.117 & 0.265 & 0.040 & 0.049 & 0.051 & 0.014 & 0.039 \\
scGPT$+$ST ridge  & 0.291 & 0.460 & 0.379 & 0.336 & 0.367 & 0.148 & 0.098 & 0.136 & 0.180 & 0.140 \\
\bottomrule
\end{tabular}}
\end{table}

\paragraph{Clustered inference.}
With gene families as the resampling unit
(Appendix~\ref{sec:methods_stats}), the full decoder is significantly
more accurate than the ridge probe in $9$ of $12$ configurations on
perturbations near training with above-median response size, and
significantly less accurate on the far slice in $8$ of $12$. The loss
from keeping only the rank-1 component is significant for scGPT$+$ST in
every cell line.

\subsection{Reliability-distance relationship across all four cell lines}
\label{sec:supp_trust_boundary_4cells}

\begin{figure}[t]
\centering
\includegraphics[width=\linewidth]{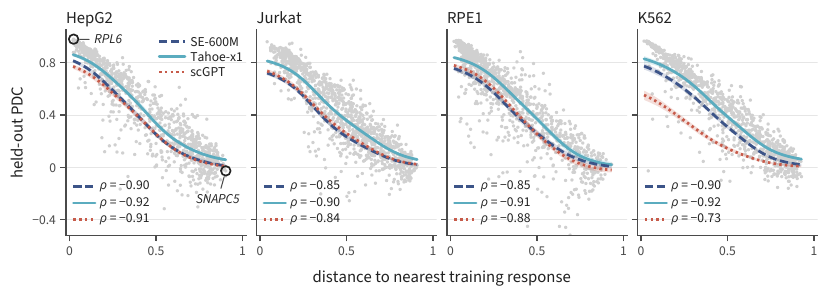}
\caption{\textbf{Reliability-distance relationship across all four cell lines.} Each
panel plots held-out per-perturbation PDC against cosine distance to
the nearest training perturbation in $2{,}000$-HVG gene-expression
space. Gray dots in the background are individual perturbations under
Tahoe-x1$+$ST. Lines are kernel-smoothed trends per encoder with $95\%$ bootstrap
bands, and the lower-left key of each panel gives the per-encoder
Spearman $\rho$. \gene{RPL6} and \gene{SNAPC5} are annotated on
the HepG2 panel as the running example
(\S\ref{sec:intro}). Every panel shows the HepG2 pattern: PDC falls
smoothly with distance and the three encoders agree. The weakest
configuration is scGPT/K562, with $\rho{=}-0.73$.}
\label{fig:supp_trust_boundary_4cells}
\end{figure}

Figure~\ref{fig:supp_trust_boundary_4cells} repeats
Figure~\ref{fig:family_interpolation}B for all four cell lines. In every
cell line PDC falls smoothly with the oracle distance, the three
encoders give curves of the same shape, and the per-encoder
Spearman~$\rho$ between the oracle distance and per-perturbation PDC is
between $-0.73$ and $-0.92$.

\subsection{Distance and position on the leading axis}
\label{sec:held_out_generalization}

\begin{figure}[h]
\centering
\includegraphics[width=\linewidth]{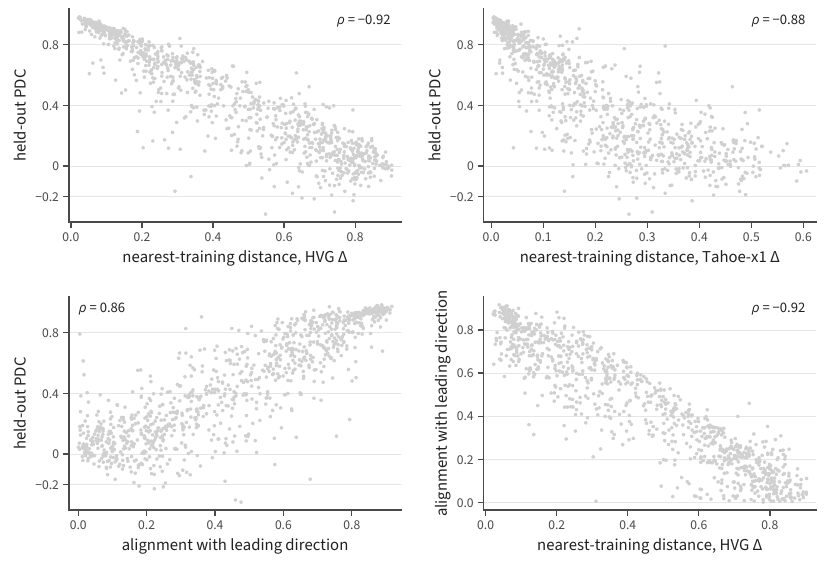}
\caption{Tahoe$+$ST on HepG2, $945$ held-out perturbations. PDC against
the distance to the nearest training response in gene space and in the
Tahoe-x1 embedding space (top), PDC against the alignment with the top
singular direction (bottom left), and the alignment against the
gene-space distance (bottom right).}
\label{fig:p1i_heldout_hepg2}
\end{figure}

A perturbation whose response lies close to the leading axis of the
responses also tends to lie close to a training response.
Table~\ref{tab:p1i_heldout} separates the two for Tahoe$+$ST, using the
distance to the nearest training response in gene space and in the
Tahoe-x1 embedding space, and the alignment of the observed response
with the top singular direction of the held-out responses
(Figure~\ref{fig:p1i_heldout_hepg2} for HepG2). Each is strongly
associated with PDC on its own: the Spearman $\rho$ is $-0.90$ to $-0.92$
for the gene-space distance and $0.74$ to $0.86$ for the alignment.
Given the alignment, the gene-space distance keeps a partial
correlation of $-0.63$ to $-0.79$ with PDC; given the distance, the
alignment keeps $0.00$ to $0.14$. In gene space, closeness to a training
response accounts for nearly all of the association. In the embedding
space the two share it more evenly (partial correlations $-0.30$ to
$-0.45$ for the distance and $0.39$ to $0.50$ for the alignment).
Appendix~\ref{sec:supp_rev_family_holdout} retrains the decoder with
whole gene families held out, so that the nearest training responses
come from other families.

\begin{table}[h]
\centering\small
\caption{Tahoe$+$ST: association of PDC with the distance to the nearest
training response in gene space ($d_{\mathrm{HVG}}$) and in the
Tahoe-x1 embedding space ($d_{\mathrm{Tahoe}}$), and with the alignment
$|\cos|$ of the observed response with the top singular direction of
the held-out responses. Spearman $\rho$ for each alone; partial Pearson
$r$ given the variable after the bar.}
\label{tab:p1i_heldout}
\resizebox{\linewidth}{!}{%
\begin{tabular}{lrrrrrrrr}
\toprule
Cell & $n_{\mathrm{test}}$ & $n_{\mathrm{train}}$ &
$\rho_{\mathrm{PDC}, d_{\mathrm{HVG}}}$ &
$\rho_{\mathrm{PDC}, d_{\mathrm{Tahoe}}}$ &
$\rho_{\mathrm{PDC}, |\cos|}$ &
$r_{\mathrm{PDC}, d_{\mathrm{HVG}} | |\cos|}$ &
$r_{\mathrm{PDC}, |\cos| | d_{\mathrm{HVG}}}$ &
$r_{\mathrm{PDC}, d_{\mathrm{Tahoe}} | |\cos|}$ \\
\midrule
HepG2  &  945 &  395 & $-0.92$ & $-0.88$ & $0.86$ & $-0.64$ & $0.06$ & $-0.38$ \\
Jurkat & 1084 &  453 & $-0.90$ & $-0.76$ & $0.74$ & $-0.76$ & $0.09$ & $-0.33$ \\
RPE1   & 1061 &  438 & $-0.91$ & $-0.86$ & $0.84$ & $-0.63$ & $0.14$ & $-0.30$ \\
K562   &  968 &  415 & $-0.92$ & $-0.81$ & $0.78$ & $-0.79$ & $0.00$ & $-0.45$ \\
\bottomrule
\end{tabular}}
\end{table}

\clearpage

\section{Extended mechanism evidence}
\label{sec:supp_mechanism_extended}

Section~\ref{sec:results_mechanism} summarizes the mechanism. Here we
give the experiments behind it for each encoder and cell line; the
interventions are defined in
Appendices~\ref{sec:methods_rank_k}--\ref{sec:methods_patching}.

\subsection{Where the perturbation enters}
\label{sec:layer_scan}

Figure~\ref{fig:hero}A shows, averaged over cell lines, the PDC after the
attention output of one layer is reduced to a random direction.
Table~\ref{tab:layer_scan} gives each cell line, together with the same
reduction to the top singular direction. Keeping the top direction
preserves most of the PDC at every layer. A random direction removes
the PDC of SE$+$ST at $L_0$ and leaves $33$--$52\%$ of that of
Tahoe$+$ST, while at $L_1$--$L_7$ it leaves at least $84\%$; the
exception is Tahoe$+$ST on K562 at $L_3$ ($76\%$). $L_0$ is the only
layer at which the kept direction matters.

Setting the $L_0$ attention output to zero brings the PDC of SE$+$ST
from $0.31$--$0.48$ to near zero on HepG2, Jurkat, and RPE1, and lowers
that of Tahoe$+$ST by $33$--$40\%$ on the same cell lines. In Tahoe$+$ST
part of the perturbation therefore reaches the output without passing
through the $L_0$ attention output (Appendix~\ref{sec:what_is_ST_doing}).

\begin{table}[h]\centering\scriptsize\setlength{\tabcolsep}{3pt}
\caption{PDC at each layer when the attention output of that layer is
reduced to one direction, as a percentage of the PDC without the
intervention. \textit{Top direction}: the top singular direction of the
layer's output. \textit{Random direction}: a random direction outside
the top $32$, with each row rescaled to its original norm. Values near
$100$ mean that the layer's output can be reduced this way with little
loss. The largest losses are for the random direction at $L_0$.}
\label{tab:layer_scan}
\begin{tabular}{llcccccccccccccccc}
\toprule
 & & \multicolumn{8}{c}{Top direction} & \multicolumn{8}{c}{Random direction} \\
\cmidrule(lr){3-10}\cmidrule(lr){11-18}
Encoder & Cell line & L0 & L1 & L2 & L3 & L4 & L5 & L6 & L7 & L0 & L1 & L2 & L3 & L4 & L5 & L6 & L7 \\
\midrule
SE+ST    & HepG2  &  99 &  98 &  97 &  98 &  97 &  99 &  99 &  99 &   4 &  99 & 100 & 101 &  97 &  99 &  95 &  98 \\
SE+ST    & Jurkat &  96 &  98 &  96 &  99 &  98 &  97 &  97 &  98 &  -3 &  95 &  94 &  96 &  96 &  94 &  96 &  97 \\
SE+ST    & RPE1   &  99 &  99 &  99 &  98 &  99 & 101 &  99 & 100 &   2 & 103 &  99 & 100 &  99 &  99 &  99 &  98 \\
SE+ST    & K562   & 102 &  99 &  98 & 100 &  99 &  99 &  99 &  99 &  -4 &  98 &  98 &  98 &  99 &  99 & 100 & 100 \\
Tahoe+ST & HepG2  &  97 &  99 &  95 &  93 &  89 &  87 &  95 & 100 &  42 &  96 &  88 &  94 &  91 &  94 &  97 &  99 \\
Tahoe+ST & Jurkat & 103 & 101 &  97 &  91 &  81 &  84 &  89 & 100 &  38 & 104 &  85 &  87 &  87 &  90 &  95 &  99 \\
Tahoe+ST & RPE1   &  97 & 100 &  95 &  89 &  89 &  93 &  93 & 100 &  52 &  97 &  93 &  88 &  92 &  95 &  98 & 100 \\
Tahoe+ST & K562   & 102 &  99 &  82 &  75 &  79 &  81 &  86 & 100 &  33 & 100 &  85 &  76 &  86 &  89 &  94 & 100 \\
\bottomrule
\end{tabular}
\end{table}

\subsection{The direction carries the signal}
\label{sec:t11_orientation}

Figure~\ref{fig:hero}B shows the size and direction interventions on
the top component $\sigma_1 u_1 v_1^\top$ of the $L_0$ output. Here we
give their ranges, the comparison behind the direction swap, and two
further replacements of the direction.

\begin{figure}[h]
\centering
\includegraphics[width=0.8\linewidth]{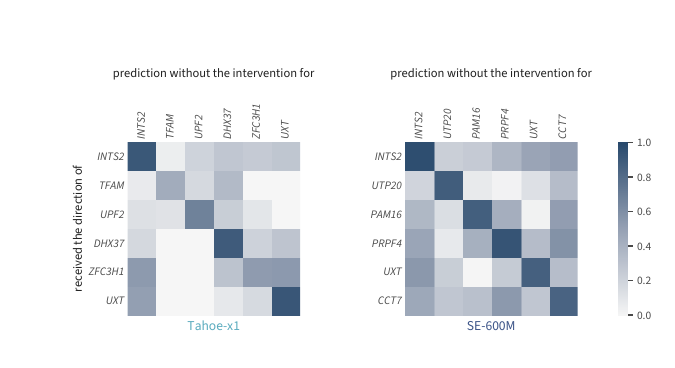}
\caption{Direction swap in K562. Each row is the prediction for a
perturbation that received the $L_0$ direction of the perturbation
named on the row, and each column is the prediction without the
intervention for the perturbation named on the column. Entries are
correlations of the responses. Six pairs per encoder that are both
predicted well and have dissimilar observed responses, chosen so that
the column perturbations differ from each other.}
\label{fig:supp_direction_swap}
\end{figure}

\paragraph{Size and direction.}
Setting $\sigma_1$ to its mean over perturbations changes mean PDC by
at most $0.003$ for Tahoe$+$ST and $0.014$ for SE$+$ST in every cell
line. Giving each perturbation the direction of another perturbation
lowers it by $0.10$--$0.20$ for Tahoe$+$ST and $0.16$--$0.23$ for
SE$+$ST. We then take pairs of perturbations $p, q$ that are both
predicted well (PDC above $0.7$) and have dissimilar observed responses
(correlation below $0.3$). After $p$ receives the direction of $q$, its
prediction is closer to the prediction for $q$ than to that for $p$ in
$69$ of $73$ pairs for Tahoe$+$ST and in all $24$ for SE$+$ST
(Figure~\ref{fig:supp_direction_swap}), and closer to the observed
response of $q$ in $66$ of $73$ and $24$ of $24$. The prediction
follows the direction it receives.

\paragraph{Direction from another encoder.}
\label{sec:cross_encoder_patching}
Giving a perturbation the direction that another encoder produces for
it, at the recipient's size, lowers PDC. The SE-600M direction lowers
Tahoe$+$ST by $0.11$--$0.29$ in every cell line. The Tahoe-x1 and scGPT
directions lower SE$+$ST by $0.21$--$0.27$ on HepG2, Jurkat, and K562;
the exception is RPE1, where the loss is below $0.03$. Replacing the
second or third component of SE$+$ST with another encoder's changes PDC
by at most $0.027$.

\begin{table}[h]
\centering\small
\caption{Change in mean PDC when the top $L_0$ direction of each
perturbation is replaced by the one the same encoder gives it in a
decoder trained on another cell line (mean over the three other cell
lines). ``Same cell line'' rows use a direction from another run on the
recipient's own cell line, as a check of the replacement itself.}
\label{tab:bundle_g_crosscell}
\begin{tabular}{llcccc}
\toprule
Encoder & Donor & HepG2 & Jurkat & RPE1 & K562 \\
\midrule
SE-600M  & other cell lines & $-0.139$ & $-0.112$ & $-0.008$ & $-0.117$ \\
         & same cell line   & $-0.013$ & $0.004$ & $-0.005$ & $-0.004$ \\
Tahoe-x1 & other cell lines & $-0.208$ & $-0.159$ & $-0.226$ & $-0.164$ \\
         & same cell line   & $-0.001$ & $-0.001$ & $0.000$ & $-0.005$ \\
scGPT    & other cell lines & $-0.057$ & $-0.073$ & $-0.007$ & $-0.082$ \\
         & same cell line   & $0.020$ & $-0.012$ & $-0.003$ & $0.036$ \\
\bottomrule
\end{tabular}
\end{table}

\paragraph{Direction from another cell line.}
\label{sec:bundle_g_wrongcell}
Table~\ref{tab:bundle_g_crosscell} replaces the direction with the one
that a decoder of the same encoder, trained on another cell line, gives
the same perturbation. Tahoe$+$ST loses $0.16$--$0.23$ in every cell
line, about as much as with the direction of another encoder. SE$+$ST
and scGPT$+$ST lose less, and almost nothing on RPE1. Except on RPE1,
the direction a decoder uses for a perturbation does not carry over to
a decoder trained on other cell lines or with another encoder.

\subsection{Rank of the \texorpdfstring{$L_0$}{L0} output and of the data}
\label{sec:rank_collapse}
\label{sec:target_rank}

Section~\ref{sec:results_mechanism} states that the $L_0$ output of
each perturbation is close to rank one. After training, its rank-1
fraction (Appendix~\ref{sec:methods_diagnostics}), averaged over
perturbations, is $0.97$--$0.996$ for SE$+$ST, $0.49$--$0.94$ for
scGPT$+$ST, and $0.82$--$0.85$ for Tahoe$+$ST. It is $0.983$ and
$0.9999$ at initialization for SE-600M and Tahoe-x1, so training lowers
it for Tahoe$+$ST by about the same amount in every cell line ($-0.160
\pm 0.012$ across three seeds on HepG2). The remaining energy of
Tahoe$+$ST is spread over several directions: after each perturbation's
mean over its $64$ cells is subtracted, the participation ratio of the
remainder is $8.1$--$9.7$, against $2.2$--$4.0$ for SE$+$ST.

The observed responses are not close to rank one
(Table~\ref{tab:target_rank}). Their participation ratio is $2.6$--$13.1$
across cell lines, and their rank-1 SVD approximation reaches a
correlation of $0.29$--$0.57$ with the responses, against $0.83$--$0.92$
for a resampled copy of the same responses. The low rank of the $L_0$
output comes from the decoder. Tahoe$+$ST, the most accurate decoder,
scores close to the rank-1 approximation on RPE1 and HepG2, between the
rank-1 and rank-4 approximations on Jurkat and K562, and $0.18$--$0.23$
below the rank-16 approximation everywhere.

\begin{table}[h]
\centering\small
\caption{Rank of the observed responses of the held-out perturbations
in each cell line ($2{,}000$ HVGs). $\mathrm{PR}$: participation ratio;
$\mathrm{PR}_{\text{cen}}$: after subtracting the mean response.
Rank $k$: mean per-perturbation correlation between the observed
responses and their top-$k$ SVD approximation. Resampled: median
correlation of each perturbation's response with itself after
resampling its cells. Last column: PDC of Tahoe$+$ST on all held-out
perturbations, from the run of Appendix~\ref{sec:nn_oracle}
(Table~\ref{tab:comparison_main} uses the perturbations scored by every
predictor).}
\label{tab:target_rank}
\begin{tabular}{lrrrrrrr}
\toprule
Cell line & $\mathrm{PR}$ & $\mathrm{PR}_{\text{cen}}$ & rank $1$ & rank $4$ & rank $16$ & resampled & Tahoe$+$ST \\
\midrule
HepG2  &  4.29 &  8.26 & 0.448 & 0.548 & 0.661 & 0.854 & 0.446 \\
Jurkat & 10.96 & 19.85 & 0.337 & 0.496 & 0.624 & 0.840 & 0.396 \\
RPE1   &  2.60 &  5.37 & 0.566 & 0.631 & 0.736 & 0.923 & 0.558 \\
K562   & 13.05 & 19.81 & 0.290 & 0.457 & 0.586 & 0.830 & 0.393 \\
\bottomrule
\end{tabular}
\end{table}

\subsection{Retrieval from the training responses}
\label{sec:nn_oracle}

Section~\ref{sec:results_mechanism} compares the decoder with retrieval
from the training responses. The retrieval oracle never runs the
decoder. For a held-out perturbation it finds the training
perturbations whose observed responses are closest, in cosine distance
over the $2{,}000$ HVGs, to the perturbation's own observed response,
and predicts a weighted average of the five nearest, with weights
$w_i \propto \exp(-d_i/\tau)$ and $\tau = \max(\mathrm{median}(d_2 -
d_1), 10^{-3})$ in each configuration. Each retrieved response is added
to the control cell the decoder used for that row, passed through the
decoder's non-negativity floor, and averaged into a pseudobulk, so the
oracle and the decoder are scored the same way.

Table~\ref{tab:nn_oracle} shows that the oracle is more accurate than
the decoder in all twelve configurations, by $0.137$ PDC on average.
Copying the nearest response instead gives a mean gain of $0.118$. The
gain is smallest for Tahoe$+$ST and largest for scGPT$+$ST on K562.

\begin{table}[h]
\centering
\caption{PDC of the five-neighbor retrieval oracle minus PDC of the
decoder, on all held-out perturbations. Positive means retrieval is
more accurate.}
\label{tab:nn_oracle}
\begin{tabular}{lcccc}
\toprule
Encoder & HepG2 & Jurkat & RPE1 & K562 \\
\midrule
SE-600M   & $+0.144$ & $+0.174$ & $+0.163$ & $+0.136$ \\
Tahoe-x1  & $+0.076$ & $+0.074$ & $+0.074$ & $+0.064$ \\
scGPT     & $+0.159$ & $+0.160$ & $+0.152$ & $+0.268$ \\
\bottomrule
\end{tabular}
\end{table}

\subsection{What the decoder adds to a linear read-out}
\label{sec:b_prime}
\label{sec:what_is_ST_doing}

Table~\ref{tab:comparison_main} includes a ridge probe on the decoder's
inputs. Here we compare the decoder with all probes of
Appendix~\ref{sec:methods_probes}, which take the decoder's three
conditioning inputs (Eq.~\eqref{eq:probe_feature}), are trained on
training perturbations, and are scored on the same genes and
perturbations as the decoder. A ridge on the control and batch inputs
alone reaches PDC $0.27$--$0.46$, the same to three decimals for the
three encoders, so what the probes know about the perturbation comes
from the decoder's perturbation embedding.

For SE$+$ST and scGPT$+$ST the probes are about as accurate as the
decoder. The ridge probe reaches mean PDC $0.377$ and $0.367$, against
$0.374$ and $0.337$, and the two-layer MLP recovers $94$--$108\%$ and
$94$--$130\%$ of the decoder's PDC (the largest ratio is scGPT$+$ST on
K562, where the decoder's PDC is small). An MLP tuned by grouped
cross-validation is not more accurate than the ridge for any encoder.

Tahoe$+$ST is the exception. Its ridge probe reaches $0.398$ against
$0.438$. The two-layer MLP recovers $80$--$84\%$ of its PDC, an MLP with
about as many parameters as the decoder's MLP layers $85$--$88\%$, and
the interaction probe $68$--$78\%$ at every rank from $0$ to $64$. The
decoder's advantage over the ridge probe is significant in all four
cell lines, also with gene families as the resampling unit
(Appendix~\ref{sec:methods_stats}).

Tahoe$+$ST is also the decoder in which part of the perturbation reaches
the output without passing through the $L_0$ attention output: setting
that output to zero leaves about two thirds of its PDC on HepG2, Jurkat,
and RPE1, while keeping only its top direction keeps $97$--$103\%$
(Appendix~\ref{sec:layer_scan}). The probes see the same inputs but not
this route.

\section{Gene programs along the leading direction}
\label{sec:supp_biology}
\label{sec:discussion_biology}

Section~\ref{sec:results_mechanism} shows that each perturbation enters
the decoder along one leading direction. Here we map this direction to
genes for Tahoe$+$ST on HepG2 ($945$ held-out perturbations, $2{,}000$
HVGs; responses are pseudobulk differences from the mean of the control
cells) and check the other cell lines.

\paragraph{The best-predicted perturbations.}
The ten held-out perturbations with the highest PDC are nine knockdowns
of large-subunit ribosomal proteins or 60S biogenesis factors and one of
the translation initiation factor \gene{EIF2B2}
(Table~\ref{tab:bio_top10}). Their observed responses share the
ribosomal-stress program described by~\citet{replogle2022mapping}:
p53 targets (\gene{MDM2}, \gene{TP53I3}, \gene{RPS27L}, \gene{FDXR})
and the lncRNAs \gene{NEAT1} and \gene{MALAT1} go up, while tubulins,
G2/M genes (\gene{MKI67}, \gene{CENPF}, \gene{BIRC5}),
replication-dependent histones and HMG proteins go down. For these ten
perturbations the predicted and observed responses have cosine
$0.90$--$0.96$ and share $13$ to $17$ of their top-20 genes with the
same sign.

\begin{table}[h]
\centering
\small
\caption{The ten held-out perturbations with the highest PDC for Tahoe$+$ST on
HepG2. Cosine: between predicted and observed responses over the
$2{,}000$ HVGs, with the observed control mean subtracted from both. Jaccard top-20: overlap of the $20$ genes with the
largest predicted and observed changes. Shared$+$sign: genes in both
top-20 lists with the same sign.}
\label{tab:bio_top10}
\begin{tabular}{lrrrrr}
\toprule
Pert & PDC & $n$ test cells & Jaccard top-20 & cos($\delta_{\text{obs}}$, $\delta_{\text{pred}}$) & shared$+$sign$/20$ \\
\midrule
RPL6   & 0.980 & 106 & 0.67 & 0.962 & 16 \\
RPL23A & 0.975 &  67 & 0.67 & 0.952 & 16 \\
RPL35A & 0.973 & 174 & 0.54 & 0.943 & 14 \\
RPL23  & 0.973 & 134 & 0.60 & 0.943 & 15 \\
RPL7   & 0.971 & 110 & 0.60 & 0.933 & 15 \\
RRS1   & 0.970 & 132 & 0.67 & 0.948 & 16 \\
EIF2B2 & 0.968 & 175 & 0.67 & 0.946 & 16 \\
RPL18  & 0.964 &  76 & 0.48 & 0.927 & 13 \\
WDR12  & 0.962 & 108 & 0.74 & 0.931 & 17 \\
NOL8   & 0.962 & 120 & 0.54 & 0.902 & 14 \\
\bottomrule
\end{tabular}
\end{table}

\paragraph{Genes along the leading direction.}
We regress each perturbation's observed response on the top right
singular vector $v_1$ of its $L_0$ output, with ridge regression and
five-fold cross-validation, and project the coefficients onto the first
principal axis of the $v_1$ vectors (Table~\ref{tab:bio_sigma1_axis}).
The largest loadings are tubulins, \gene{STMN1}, \gene{H2AFZ},
\gene{HMGB1}/\gene{HMGB2}, \gene{CENPF}, and \gene{MKI67} on one side
and \gene{MALAT1}, \gene{NEAT1}, \gene{MDM2}, and \gene{RHOC} on the
other, so the direction runs from proliferating to arrested,
p53-active cells. It is a linear summary of the decoder: across genes,
the cross-validated $R^2$ is $0.12$ at the median and at most $0.62$
(\gene{STMN1}).

\begin{table}[h]
\centering
\small
\caption{Genes with the largest loadings on the leading direction for
Tahoe$+$ST on HepG2. The sign is that of the first principal axis of
the $v_1$ vectors and has no biological orientation.}
\label{tab:bio_sigma1_axis}
\begin{tabular}{lrlrlrlr}
\toprule
gene & loading & gene & loading & gene & loading & gene & loading \\
\midrule
TUBA1B & $1.06$ & STMN1 & $1.02$ & H2AFZ & $0.90$ & MALAT1 & $-0.75$ \\
HMGB2 & $0.73$ & TUBB & $0.72$ & HMGB1 & $0.70$ & CENPF & $0.67$ \\
NEAT1 & $-0.67$ & HIST1H4C & $0.63$ & PCLAF & $0.62$ & MDM2 & $-0.62$ \\
MKI67 & $0.61$ & HMGN2 & $0.59$ & BIRC5 & $0.58$ & NASP & $0.58$ \\
CALR & $0.57$ & MCM3 & $0.56$ & PTMA & $0.54$ & HSP90B1 & $0.52$ \\
SCD & $0.52$ & ACTG1 & $0.51$ & TOP2A & $0.51$ & TPX2 & $0.49$ \\
DBI & $0.48$ & ACAT2 & $0.48$ & PTTG1 & $0.48$ & RHOC & $-0.48$ \\
TK1 & $0.47$ & FADS1 & $0.46$ & & & & \\
\bottomrule
\end{tabular}
\end{table}

\paragraph{Other cell lines.}
Table~\ref{tab:bio_sigma1_hallmark} gives the top Hallmark category of
the $30$ genes with the largest loadings on the first principal axis of
each cell line's observed responses, which has cosine $0.91$--$0.99$
with the leading direction: E2F targets in HepG2 and RPE1, Myc targets
in Jurkat, and mTORC1 signaling, tied with hypoxia, in K562. In every
cell line the category comes from the proliferation side of the axis,
except one of the eight mTORC1 genes and two of the six hypoxia genes in
K562; the arrested, p53 side is identified in HepG2. The axis is close to the mean
response of all held-out perturbations (cosine $0.98$--$1.00$), that
is, to the shift that perturbed cells share relative to control cells,
which \citet{vinastorne2026systema} call systematic variation. In the
Replogle RPE1 screen they find $46\%$ of perturbed and $25\%$ of control
cells in G1, which they attribute to arrest of p53-positive RPE1 cells
after perturbation-induced chromosomal instability, and a much smaller
cell-cycle shift in p53-negative K562 cells. In K562 and Jurkat a
perturbation's position on the axis also follows its change in total
UMIs per cell: the Pearson correlation with the change in mean log UMI
count is $0.53$ and $0.74$, against $0.22$ in HepG2 and $0.03$ in RPE1.
It stays at $0.49$ and $0.68$ when pseudobulks sum counts before
normalizing, so it reflects a change in total RNA per cell rather than
the log transform. In K562 and Jurkat the axis therefore combines the
proliferation program with the change in total RNA.

\begin{table}[h]
\centering
\small
\caption{Hallmark category with the smallest adjusted $p$-value (MSigDB Hallmark
2020, Benjamini--Hochberg, background the $2{,}000$ HVGs) among the
$30$ genes with the largest absolute loadings on the first right
singular vector of each cell line's matrix of observed responses.}
\label{tab:bio_sigma1_hallmark}
\begin{tabular}{lll}
\toprule
Cell line & Top-FDR Hallmark category & $p_{\mathrm{adj}}$ \\
\midrule
HepG2  & E2F-Targets       & $4.78 \times 10^{-8}$ \\
RPE1   & E2F-Targets       & $2.14 \times 10^{-9}$ \\
Jurkat & Myc-Targets-V1    & $1.04 \times 10^{-5}$ \\
K562   & mTORC1-Signaling  & $2.67 \times 10^{-4}$ \\
       & Hypoxia (tied)    & $2.67 \times 10^{-4}$ \\
\bottomrule
\end{tabular}
\end{table}

\paragraph{Gene-level agreement.}
Over all $945$ held-out perturbations the cosine between predicted and
observed responses has median $0.26$ (Table~\ref{tab:bio_full_dist})
and rises with PDC (Spearman $0.91$; Table~\ref{tab:bio_decile}). In the
three lowest PDC deciles its median is below $0.06$: for these
perturbations, mostly with weak responses, the prediction does not
point along the observed response. The ten perturbations with the
lowest cosine all have weak responses (norm at most $4.8$) and slightly
negative cosines (for example \gene{GFOD2} $-0.24$ and \gene{DHX29}
$-0.19$).

\begin{table}[h]
\centering
\small
\caption{Agreement between predicted and observed responses over all $945$
held-out perturbations (Tahoe$+$ST, HepG2), by percentile.}
\label{tab:bio_full_dist}
\begin{tabular}{lrrr}
\toprule
percentile & cosine & Jaccard top-20 & shared+sign / 20 \\
\midrule
$p10$    & $-0.034$ & $0.000$ & $0$ \\
$p25$    & $0.056$ & $0.026$ & $0$ \\
$\mathbf{median}$ & $\mathbf{0.257}$ & $\mathbf{0.081}$ & $\mathbf{3}$ \\
$p75$    & $0.648$ & $0.290$ & $9$ \\
$p90$    & $0.858$ & $0.575$ & $15$ \\
\bottomrule
\end{tabular}
\end{table}

\begin{table}[h]
\centering
\small
\caption{Agreement between predicted and observed responses by PDC decile
($\approx 95$ perturbations each; decile $1$ has the lowest PDC).}
\label{tab:bio_decile}
\begin{tabular}{rrrrr}
\toprule
PDC decile & PDC med & cosine $p25$ & cosine median & Jaccard med \\
\midrule
1  & $-0.045$ & $-0.060$ & $-0.007$ & $0.026$ \\
2  & $0.056$ & $-0.033$ & $0.027$ & $0.026$ \\
3  & $0.142$ & $-0.008$ & $0.058$ & $0.026$ \\
4  & $0.239$ & $0.061$ & $0.122$ & $0.026$ \\
5  & $0.372$ & $0.132$ & $0.186$ & $0.053$ \\
6  & $0.520$ & $0.259$ & $0.353$ & $0.111$ \\
7  & $0.655$ & $0.354$ & $0.475$ & $0.176$ \\
8  & $0.771$ & $0.561$ & $0.651$ & $0.290$ \\
9  & $0.879$ & $0.740$ & $0.812$ & $0.481$ \\
10 & $0.934$ & $0.866$ & $0.890$ & $0.667$ \\
\bottomrule
\end{tabular}
\end{table}

\begin{figure}[t]
\centering
\includegraphics[width=\linewidth]{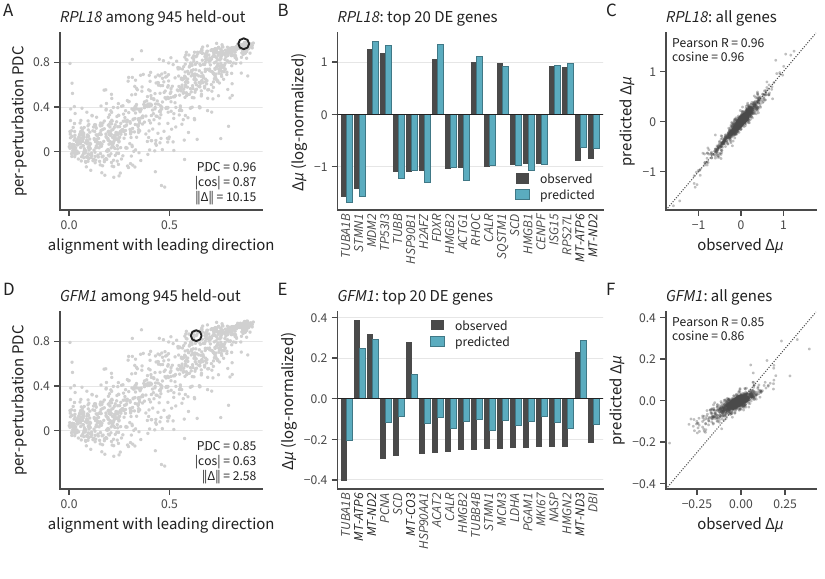}
\caption{\gene{RPL18} (top) and \gene{GFM1} (bottom) under Tahoe$+$ST on
HepG2. (A, D) Cosine of each held-out perturbation's response with the
leading direction against its PDC, with the perturbation marked.
(B, E) The $20$ genes with the largest observed change, observed (grey)
and predicted; mitochondrially encoded transcripts are labelled.
(C, F) Predicted against observed change for all $2{,}000$ HVGs, each
relative to its own control mean, with their cosine and Pearson $R$
(the PDC).}
\label{fig:p0f_walkthrough}
\end{figure}

\paragraph{A well-predicted perturbation off the direction.}
A high PDC does not imply agreement at the level of genes.
\gene{GFM1}, a mitochondrial translation factor, has PDC $0.85$, and
its $20$ most changed genes include four mitochondrially encoded
transcripts (\gene{MT-ATP6}, \gene{MT-ND2}, \gene{MT-CO3},
\gene{MT-ND3}) next to the proliferation genes of the leading direction
(Figure~\ref{fig:p0f_walkthrough}). The prediction contains three of
the four with smaller changes, and eight of its other top-20 genes are
proliferation genes. Predicted and observed top-20 lists share $7$
genes with the same sign, and their cosine, defined as in
Table~\ref{tab:bio_top10}, is $0.61$. \gene{RPL18}, whose response lies closer to the leading
direction, reaches $0.93$ and shares $13$ genes.

\begin{figure}[h]
\centering
\includegraphics[width=\linewidth]{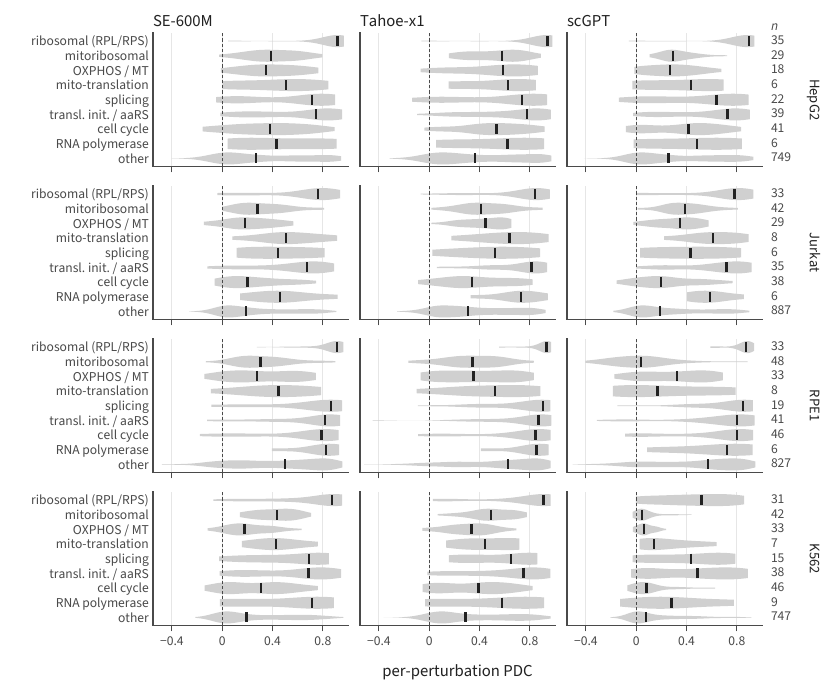}
\caption{Per-perturbation PDC by gene family for the three encoders (columns)
in the four cell lines (rows). Violins show the distribution and ticks
the median; $n$ is the number of perturbations in each family.}
\label{fig:p0e_violins}
\end{figure}

\paragraph{Gene families.}
We assign each held-out perturbation to the first matching family:
cytoplasmic ribosomal (\gene{RPL*}, \gene{RPS*}, \gene{RACK*}); mitoribosomal
(\gene{MRPL*}, \gene{MRPS*}); OXPHOS (\gene{MT-*}, \gene{NDUF*},
\gene{COX*}, \gene{ATP5*}, \gene{ATP6*}, \gene{UQCR*}, \gene{SDH[A-D]});
mitochondrial translation (\gene{GFM1/2}, \gene{TSFM}, \gene{TFAM},
\gene{TFB1M}, \gene{TUFM}, \gene{OPA1}, \gene{LONP1/2},
\gene{MRPP1-4}, \gene{PNPT1}, \gene{FASTKD5}); splicing (\gene{SNRP*},
\gene{SF3*}, \gene{U2AF*}, \gene{PRPF*}, \gene{RBM*}); translation
initiation and tRNA synthetases (\gene{EIF*} and the aaRS genes); cell
cycle and replication (\gene{CDC*}, \gene{CDK*}, \gene{CCN*},
\gene{MCM*}, \gene{CENP*}, \gene{BUB*}, \gene{AURK*}, \gene{KIF*},
\gene{PLK*}, \gene{CHK*}, \gene{RAD*}, \gene{BRCA*}, \gene{TOP*},
\gene{RFC*}, \gene{CHEK*}, \gene{ORC*}, \gene{WEE*}, and \gene{POL*}
other than \gene{POLR*}); RNA polymerase (\gene{POLR*}); and other. In all twelve configurations the
cytoplasmic ribosomal family has the highest median PDC, and the three
mitochondrial families have lower medians than both the cytoplasmic
ribosomal and the translation-initiation families
(Figure~\ref{fig:p0e_violins}). For Tahoe$+$ST the median PDC is
$0.84$--$0.94$ for cytoplasmic ribosomal perturbations and
$0.36$--$0.58$ for the three mitochondrial families together. In HepG2,
Jurkat, and K562 the mitochondrial responses are about as large as those
of the other perturbations; in RPE1 they are smaller.

\clearpage

\section{Scores that separate perturbations}
\label{sec:dir4_sigma1_eval_main}
\label{sec:dir4_sigma1_eval}

PDC is a correlation over genes, so a prediction that reproduces the
shared response of Appendix~\ref{sec:supp_biology} scores well on every
perturbation whose response lies along it, without telling perturbations
apart. We add two scores that measure this. For each held-out
perturbation we take the cells' responses (each cell minus the mean of
the observed control cells, over the $2{,}000$ HVGs) and their top right
singular vector, the gene-space direction of the response, for the
predicted and for the observed cells. \emph{Direction recovery} is the
absolute cosine between the predicted and observed directions, averaged
over perturbations. \emph{Identity retrieval} asks whether the predicted
directions keep which perturbations resemble each other: for each
held-out perturbation we take the other held-out perturbation whose
predicted direction is closest to its predicted direction, and the one
whose observed direction is closest to its observed direction, and
report the fraction of perturbations for which the two are the same.

Tahoe$+$ST recovers the direction best in every cell line, by
$0.026$--$0.068$ over SE$+$ST, and reaches $0.874$ on RPE1. Identity
retrieval is low for all three decoders: $0.034$--$0.054$ for SE$+$ST,
$0.050$--$0.074$ for Tahoe$+$ST, and $0.013$--$0.046$ for scGPT$+$ST.
Chance level is $1/(N-1)$, about $0.001$. An oracle that predicts the
response of the training perturbation whose observed response is
closest to the perturbation's own gives $0.201$, $0.109$, and $0.123$
on HepG2, Jurkat, and RPE1. The decoders recover the direction of each
response well but keep little of which perturbations resemble each
other, in line with Appendices~\ref{sec:t11_orientation}
and~\ref{sec:supp_biology}.

\section{Predicted-NN-distance as a confidence score}
\label{sec:supp_predicted_nn_followups}

Section~\ref{sec:results_deployable} evaluates predicted-NN-distance as a
confidence score. Here we compare it with other scores, test it among
predictions of similar size, and give the details of the pooled
threshold and of Learn-then-Test.

\subsection{Design of the score}
\label{sec:supp_score_design}

\paragraph{Other scores.}
\label{sec:supp_score_baselines}

Table~\ref{tab:supp_score_baselines} compares predicted-NN-distance with
four other scores computed from the prediction and the training
responses. Averaged over the $12$ configurations, its Spearman $|\rho|$
with PDC is $0.806$, against $0.784$ for the mean distance to the ten
nearest training responses, $0.555$ for the Euclidean distance, and
$0.664$ for the alignment with the leading axis of the training
responses (paired Wilcoxon $p < 0.01$ for all three). The norm of the
predicted response reaches $0.725$. It is higher than predicted-NN-distance
in the three RPE1 configurations, and the difference over the $12$
configurations is not significant ($p = 0.055$);
Appendix~\ref{sec:supp_pred_nn_magnitude_stratified} compares the two
among predictions of similar norm.

\begin{table}[h]
\centering\small
\caption{Spearman $|\rho|$ between each score and PDC on the held-out
perturbations. (a) predicted-NN-distance (cosine, nearest training
response); (b) norm of the predicted response; (c) mean cosine distance
to the ten nearest training responses; (d) Euclidean distance to the
nearest training response; (e) alignment of the predicted response with
the leading axis of the training responses.}
\label{tab:supp_score_baselines}
\begin{tabular}{llrrrrr}
\toprule
Encoder & Cell line & (a) & (b) & (c) & (d) & (e) \\
\midrule
SE    & HepG2  & 0.844 & 0.769 & 0.827 & 0.322 & 0.744 \\
SE    & Jurkat & 0.832 & 0.692 & 0.819 & 0.498 & 0.652 \\
SE    & RPE1   & 0.817 & 0.822 & 0.797 & 0.707 & 0.712 \\
SE    & K562   & 0.878 & 0.751 & 0.870 & 0.511 & 0.719 \\
Tahoe & HepG2  & 0.840 & 0.823 & 0.808 & 0.585 & 0.702 \\
Tahoe & Jurkat & 0.850 & 0.828 & 0.822 & 0.688 & 0.662 \\
Tahoe & RPE1   & 0.724 & 0.798 & 0.662 & 0.703 & 0.527 \\
Tahoe & K562   & 0.862 & 0.833 & 0.833 & 0.670 & 0.626 \\
scGPT & HepG2  & 0.830 & 0.776 & 0.819 & 0.485 & 0.754 \\
scGPT & Jurkat & 0.852 & 0.758 & 0.840 & 0.600 & 0.695 \\
scGPT & RPE1   & 0.559 & 0.730 & 0.524 & 0.669 & 0.404 \\
scGPT & K562   & 0.785 & 0.115 & 0.782 & 0.222 & 0.765 \\
\midrule
\multicolumn{2}{l}{Mean} & 0.806 & 0.725 & 0.784 & 0.555 & 0.664 \\
\bottomrule
\end{tabular}
\end{table}

\label{sec:supp_rev_hvg}

\paragraph{HVGs and the full transcriptome.}
The $2{,}000$-gene panel is fixed by the ST decoder's output layer, so
scoring all transcripts would require retraining every model. On the
observed responses, the leading axis of the $2{,}000$-HVG space and of
the full $6{,}546$-gene space have cosine $\geq 0.99$ on three of four
cell lines and $0.89$ on K562, and the association between the
oracle distance and PDC changes by less than $0.02$ in all $12$ configurations
($|\rho| = 0.873$ on the full panel against $0.876$ on HVGs). scGPT
without the ST decoder outputs all $6{,}546$ genes; scored on every
gene, its mean PDC changes by $-0.053$ to $+0.017$ relative to the HVG
panel. On the full transcriptome the leading axis explains a smaller
share of the variance.

\paragraph{Aggregating distances.}
Besides the minimum distance, we tested $24$ aggregators of the
distances from a prediction to the training responses: means of the
bottom $k$ or bottom percentile, percentiles, soft-min at several
temperatures, and neighbor counts within a radius. None is significantly
better than the minimum; the mean of the bottom $1\%$ lowers $|\rho|$ by $0.010$ and
improves on the minimum in no configuration. The number of training
neighbors within cosine $0.15$ carries a small additional signal
(mean partial $\rho$ with PDC given the minimum distance $+0.069$,
positive in $10$ of $12$), but $32$--$96\%$ of held-out perturbations have
no neighbor at that radius.

\paragraph{Other distance spaces.}
\label{sec:supp_pred_nn_alt_space}

We also computed the distance in two other spaces. In the space of
perturbation embeddings, the cosine distance between a held-out
perturbation's column of the decoder's perturbation encoder and the
closest training column gives a weaker score than the gene-space distance
in all $12$ configurations ($|\rho|$ lower by $0.33$ to $0.74$, $0.48$
on average). Along the decoder's leading $L_0$ coordinate, the gap to
the nearest training perturbation has Spearman $\rho$ of $+0.20$ to
$+0.45$ with PDC, and the gene-space distance keeps a partial
correlation of $-0.87$ to $-0.48$ with PDC given this gap.

\subsection{Beyond the size of the prediction}
\label{sec:supp_pred_nn_magnitude_stratified}

\begin{figure}[t]
\centering
\includegraphics[width=\linewidth]{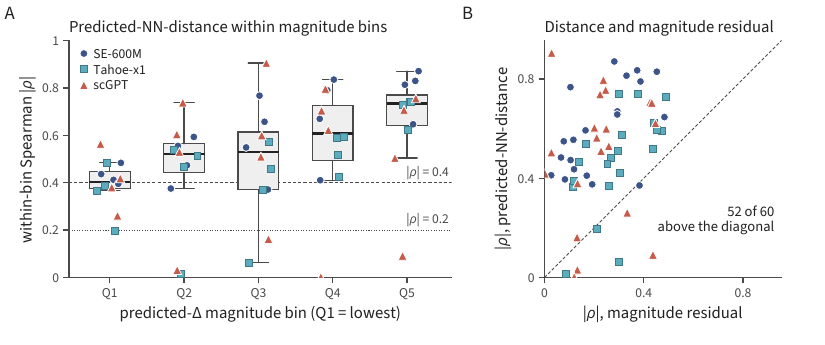}
\caption{(A) Within-bin $|\rho|$ between predicted-NN-distance and PDC in five
bins of predicted norm, one marker per configuration; boxes summarize
the $12$ configurations; dotted and dashed lines at $0.2$ and $0.4$.
(B) For each configuration and bin, $|\rho|$ of predicted-NN-distance
against $|\rho|$ of the remaining variation in predicted norm.}
\label{fig:supp_exp1_per_bin}
\end{figure}

Predictions with a large predicted response lie closer to a training
response: the Spearman $\rho$ between predicted-NN-distance and the
norm of the predicted response is $-0.72$ to $-0.89$ in $11$ of the $12$
configurations ($+0.27$ for scGPT$+$ST on K562). We therefore split each
configuration's held-out perturbations into five equal-count bins of
predicted norm and compute, within each bin, the Spearman $\rho$
between predicted-NN-distance and PDC
(Tables~\ref{tab:supp_exp1_per_config} and~\ref{tab:supp_exp1_aggregate},
Figure~\ref{fig:supp_exp1_per_bin}).

Within bins the median $|\rho|$ over configurations is $0.40$ in the bin
of smallest predictions and $0.73$ in the bin of largest. In $10$ of the
$12$ configurations $|\rho|$ is at least $0.2$ in three or more bins;
the exceptions are Tahoe$+$ST and scGPT$+$ST on RPE1, and finer bins
($7$ or $10$) give the same pattern. Within the same bins
predicted-NN-distance is more strongly associated with PDC than the
remaining variation in predicted norm in $52$ of the $60$ bins (paired
Wilcoxon over configurations, $p = 0.0012$). Regressing the predicted
norm out of both the score and PDC leaves a median partial $|\rho|$ of
$0.58$, against $0.84$ without it.

\begin{figure}[t]
\centering
\includegraphics[width=\linewidth]{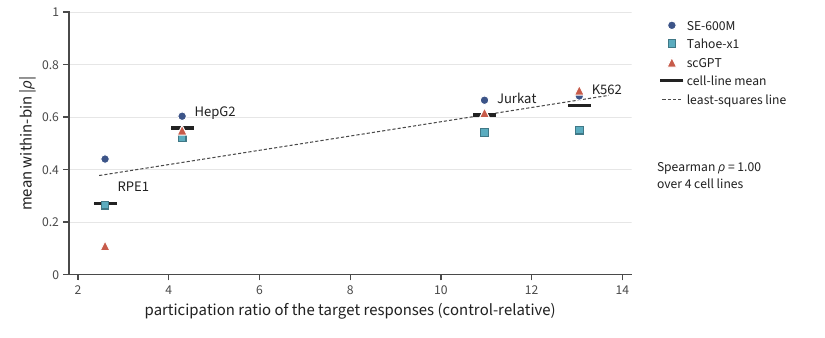}
\caption{Mean within-bin $|\rho|$ against the participation ratio of the observed
responses in each cell line, one marker per encoder; bars give the
mean over encoders.}
\label{fig:supp_exp1_step7}
\end{figure}

The loss from binning is largest where the observed responses lie
closest to one direction (Table~\ref{tab:per_cell_stratified_pr},
Figure~\ref{fig:supp_exp1_step7}). In RPE1, with participation ratio
$2.6$, the score keeps $39\%$ of its unbinned $|\rho|$, against
$66$--$77\%$ in the other cell lines. The oracle distance keeps a partial
$|\rho|$ of $0.81$ (median) and $0.86$ for Tahoe$+$ST on RPE1. The
observed responses in RPE1 therefore still carry this information; the
predictions there collapse toward the leading direction and lose it.

\begin{table}[h]
\centering\small
\caption{Spearman $\rho$ between predicted-NN-distance and PDC on all held-out
perturbations (unbinned) and within five equal-count bins of the norm
of the predicted response (Q1 smallest). MOD: at least $3$ of $5$ bins
with $|\rho| \geq 0.2$; STR: at least $4$ of $5$ bins with
$|\rho| \geq 0.4$.}
\label{tab:supp_exp1_per_config}
\resizebox{\linewidth}{!}{%
\begin{tabular}{llrrrrrrrcc}
\toprule
Encoder & Cell line & $n_{\text{test}}$ & unbin. $\rho$ & Q1 & Q2 & Q3 & Q4 & Q5 & MOD & STR \\
\midrule
SE       & HepG2  &   945 & $-0.844$ & $-0.412$ & $-0.555$ & $-0.549$ & $-0.669$ & $-0.830$ & \checkmark & \checkmark \\
SE       & Jurkat & 1084 & $-0.832$ & $-0.484$ & $-0.473$ & $-0.657$ & $-0.835$ & $-0.871$ & \checkmark & \checkmark \\
SE       & RPE1   & 1061 & $-0.817$ & $-0.395$ & $-0.375$ & $-0.371$ & $-0.410$ & $-0.647$ & \checkmark & --- \\
SE       & K562   &   968 & $-0.878$ & $-0.436$ & $-0.593$ & $-0.767$ & $-0.790$ & $-0.813$ & \checkmark & \checkmark \\
Tahoe-x1 & HepG2  &   945 & $-0.840$ & $-0.367$ & $-0.536$ & $-0.368$ & $-0.589$ & $-0.739$ & \checkmark & --- \\
Tahoe-x1 & Jurkat & 1084 & $-0.850$ & $-0.483$ & $-0.512$ & $-0.458$ & $-0.517$ & $-0.741$ & \checkmark & \checkmark \\
Tahoe-x1 & RPE1   & 1061 & $-0.724$ & $-0.196$ & $-0.016$ & $0.063$ & $-0.423$ & $-0.623$ & --- & --- \\
Tahoe-x1 & K562   &   968 & $-0.862$ & $-0.388$ & $-0.466$ & $-0.573$ & $-0.594$ & $-0.729$ & \checkmark & \checkmark \\
scGPT    & HepG2  &   945 & $-0.830$ & $-0.380$ & $-0.530$ & $-0.510$ & $-0.622$ & $-0.707$ & \checkmark & \checkmark \\
scGPT    & Jurkat & 1084 & $-0.852$ & $-0.419$ & $-0.604$ & $-0.600$ & $-0.704$ & $-0.756$ & \checkmark & \checkmark \\
scGPT    & RPE1   & 1061 & $-0.559$ & $-0.261$ & $0.031$ & $0.162$ & $0.002$ & $-0.091$ & --- & --- \\
scGPT    & K562   &   968 & $-0.785$ & $-0.564$ & $-0.738$ & $-0.905$ & $-0.796$ & $-0.504$ & \checkmark & \checkmark \\
\bottomrule
\end{tabular}}
\end{table}

\begin{table}[h]
\centering\small
\caption{Within-bin $|\rho|$ over the $12$ configurations: median,
interquartile range, and the number of configurations above $0.2$ and
$0.4$.}
\label{tab:supp_exp1_aggregate}
\begin{tabular}{rrrrr}
\toprule
Bin & Cross-config median $|\rho|$ & IQR ($25\% / 75\%$) & $\#|\rho| \geq 0.2$ & $\#|\rho| \geq 0.4$ \\
\midrule
Q1 & $0.404$ & $0.377 / 0.448$ & $11/12$ & $6/12$ \\
Q2 & $0.521$ & $0.443 / 0.565$ & $10/12$ & $9/12$ \\
Q3 & $0.529$ & $0.370 / 0.614$ & $10/12$ & $8/12$ \\
Q4 & $0.608$ & $0.494 / 0.726$ & $11/12$ & $11/12$ \\
Q5 & $0.734$ & $0.641 / 0.770$ & $11/12$ & $11/12$ \\
\bottomrule
\end{tabular}
\end{table}

\begin{table}[h]
\centering\small
\caption{Mean within-bin $|\rho|$ (over the three encoders and five bins) and
mean unbinned $|\rho|$ in each cell line, their ratio, and the
participation ratio of the observed responses
(Table~\ref{tab:target_rank}).}
\label{tab:per_cell_stratified_pr}
\begin{tabular}{lcccc}
\toprule
Cell line & target $\mathrm{PR}_{\text{ctrl-rel}}$ & mean stratified $|\rho|$ & mean unbinned $|\rho|$ & attenuation ratio \\
\midrule
HepG2  & $4.29$  & $0.558$ & $0.838$ & $0.665$ \\
Jurkat & $10.96$ & $0.608$ & $0.845$ & $0.719$ \\
RPE1   & $2.60$  & $0.271$ & $0.700$ & $0.387$ \\
K562   & $13.05$ & $0.644$ & $0.842$ & $0.765$ \\
\bottomrule
\end{tabular}
\end{table}

\subsection{Pooled threshold}
\label{sec:supp_pooled_tau}

\paragraph{Selection and leave-one-out refits.}

The pooled threshold is chosen on a grid of $50$ values of $\tau$
between $0.048$ and $0.82$. For each value we compute, in each of the
$12$ configurations, the mean PDC of the accepted predictions above that
cell line's rank-$1$ SVD baseline, and we keep the value with the largest
mean over configurations among those at which every configuration
accepts at least five predictions. This gives $\tau = 0.0957$. Over
$5{,}000$ bootstrap resamples that draw configurations and, within each,
perturbations with replacement, with the grid rebuilt in each resample,
the selected $\tau$ has mean $0.099$ and $95\%$ interval
$[0.073, 0.116]$. Refitting with one cell line left out (four folds) or
one encoder left out (three folds) selects $\tau$ between $0.092$ and
$0.107$. Applied to the left-out cell line or encoder, the refitted
threshold changes the mean PDC of accepted predictions above the
baseline by at most $0.011$ (cell lines) and $0.019$ (encoders), and by
more than $0.05$ in none of the $11$ configurations that accept at
least five predictions in every refit. Because the pooled threshold is
chosen on the same held-out predictions that
Figure~\ref{fig:universal_threshold}B reports, these leave-one-out refits
measure how much this choice inflates the reported accuracy.

\paragraph{Transfer between configurations.}
\label{sec:supp_rev_transfer}

Figure~\ref{fig:supp_transfer_12x12} applies the per-configuration
threshold of Appendix~\ref{sec:methods_thresholds}, fitted on each source
configuration (rows), to each target configuration (columns), scored as
the mean PDC of accepted predictions above the target's rank-$1$ SVD
baseline. Because the per-configuration thresholds differ, a threshold
taken from another configuration accepts a different share of the
target's predictions, so the entries vary in both directions around the
diagonal. One pair (scGPT/RPE1 as source, SE/Jurkat as target) accepted
fewer than five predictions and is masked.

\begin{figure}[h]
\centering
\includegraphics[width=0.9\linewidth]{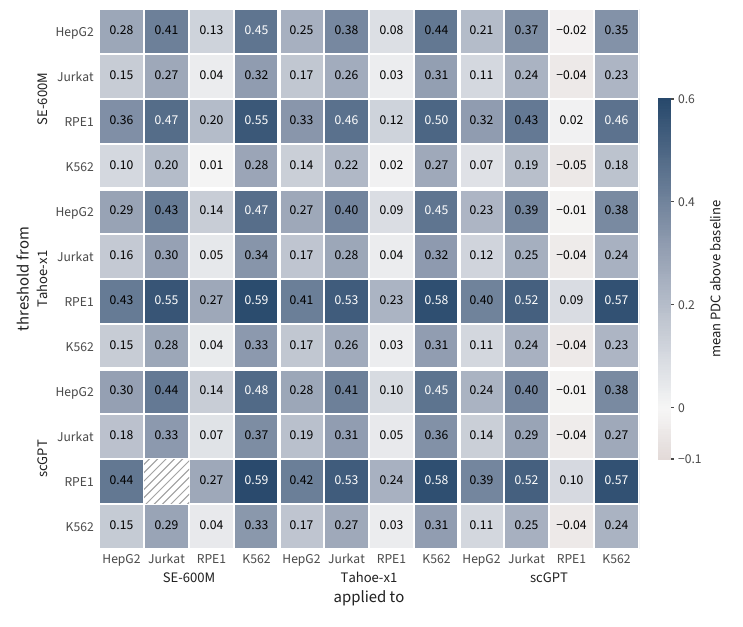}
\caption{Transfer of the per-configuration threshold between all
$12 \times 12$ source--target pairs. Entry: mean PDC above the rank-$1$
SVD baseline among predictions accepted on the target (columns) using
the threshold fitted on the source (rows). Hatched: fewer than five
accepted predictions. White rules separate the encoder blocks.}
\label{fig:supp_transfer_12x12}
\end{figure}

\paragraph{Coverage.}
\label{sec:supp_pred_nn_coverage}

\begin{figure}[t]
\centering
\includegraphics[width=\linewidth]{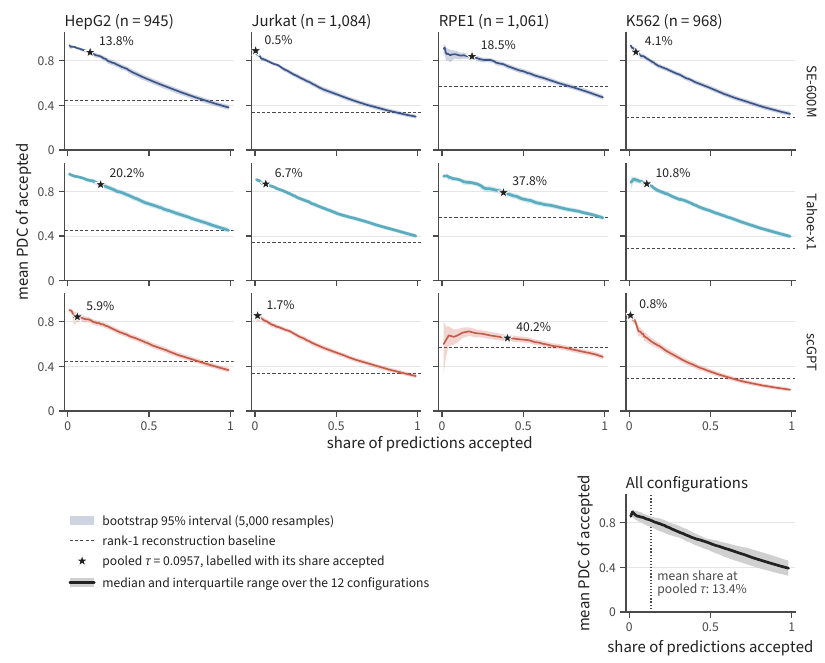}
\caption{Mean PDC of the accepted predictions against the share accepted, as
the threshold on predicted-NN-distance is raised (rows: encoders;
columns: cell lines). Shaded: bootstrap $95\%$ interval ($5{,}000$
resamples); dashed: rank-1 SVD baseline; star: the pooled threshold,
labeled with its share accepted. Lower right: median and interquartile
range over the $12$ configurations.}
\label{fig:supp_coverage_pareto}
\end{figure}

Figure~\ref{fig:supp_coverage_pareto} gives, for each configuration,
the mean PDC of the accepted predictions as the threshold is raised,
the per-configuration version of Figure~\ref{fig:universal_threshold}A.
The pooled threshold accepts $13.4\%$ of the predictions on average,
with mean PDC $0.43$ above the rank-1 SVD baseline. Over
configurations, the median mean PDC of the accepted predictions is
$0.86$ at $5\%$ of predictions accepted, $0.72$ at $30\%$, and $0.61$ at
$50\%$, still $0.17$ above the baseline.

\subsection{Learn-then-Test threshold}
\label{sec:supp_rev_ltt}

Table~\ref{tab:supp_ltt} compares Learn-then-Test with a decile rule
over $200$ random splits into calibration and deployment halves. For
each configuration it reports the mean share of all predictions
accepted and the realized mean PDC of the accepted predictions.
Learn-then-Test uses target $0.30$, $\delta = 0.10$, a $201$-point
grid, and Bonferroni-simultaneous one-sided Hoeffding bounds. The
decile rule sorts the calibration perturbations by predicted-NN-distance
into ten equal-count deciles, computes a one-sided $95\%$
percentile-bootstrap lower bound ($1{,}000$ resamples) on the mean PDC
of each decile, and accepts the largest contiguous prefix of
lowest-distance deciles whose bounds are at least the target $0.30$;
$\tau$ is the largest calibration distance in that prefix. A split that
Learn-then-Test does not certify counts as accepting nothing. No
certified split in any configuration had realized mean PDC below
$0.30$.

The guarantee holds when the calibration and deployment perturbations
are drawn from the same population, as in these random splits. To check
a split that breaks this condition, we calibrate on the half of the
held-out perturbations nearest the training responses (by oracle
distance) and deploy on the far half. The realized mean PDC of the
accepted predictions then falls below $0.30$ in $11$ of the $12$
configurations.

\begin{table}[h]
\centering\small
\caption{Learn-then-Test and the decile rule over $200$ random
calibration/deployment splits: share of predictions accepted and mean
PDC of accepted predictions.}
\label{tab:supp_ltt}
\begin{tabular}{llcccc}
\toprule
 & & \multicolumn{2}{c}{Learn-then-Test} & \multicolumn{2}{c}{Decile rule} \\
\cmidrule(lr){3-4}\cmidrule(lr){5-6}
Encoder & Cell line & Accepted & PDC & Accepted & PDC \\
\midrule
SE    & HepG2  & 0.696 & 0.520 & 0.465 & 0.654 \\
SE    & Jurkat & 0.380 & 0.574 & 0.342 & 0.600 \\
SE    & K562   & 0.506 & 0.553 & 0.419 & 0.603 \\
SE    & RPE1   & 0.987 & 0.475 & 0.612 & 0.651 \\
Tahoe & HepG2  & 0.897 & 0.494 & 0.584 & 0.646 \\
Tahoe & Jurkat & 0.759 & 0.493 & 0.535 & 0.590 \\
Tahoe & K562   & 0.725 & 0.510 & 0.513 & 0.618 \\
Tahoe & RPE1   & 1.000 & 0.558 & 0.826 & 0.622 \\
scGPT & HepG2  & 0.630 & 0.532 & 0.459 & 0.628 \\
scGPT & Jurkat & 0.447 & 0.552 & 0.387 & 0.587 \\
scGPT & K562   & 0.000 & not certified & 0.213 & 0.557 \\
scGPT & RPE1   & 0.998 & 0.481 & 0.690 & 0.583 \\
\midrule
\multicolumn{2}{l}{Mean} & 0.669 & & 0.504 & \\
\bottomrule
\end{tabular}
\end{table}

\clearpage

\section{Tests outside the audited panel}
\label{sec:supp_generalization_tests}

Section~\ref{sec:results_generalization} tests the score outside the
audited setting. Here we give the details for TxPert, for chemical
perturbations in Tahoe-100M, for single-cell scoring and other context
sizes, and for retraining with whole gene families held out.

\subsection{TxPert}
\label{sec:supp_rev_txpert}

We ran the released TxPert~\citep{wenkel2026txpert} GAT and Exphormer
checkpoints for unseen perturbations on their $272$ held-out Replogle
K562 perturbations. We intervened on the $64$-dimensional perturbation
latent that enters TxPert's two-layer MLP decoder, the closest analogue
of the ST decoder's $L_0$. PDC follows TxPert's evaluation: both
responses are centred on the observed batch-matched control, over
TxPert's gene set. The projection basis is the SVD of all latent rows,
including test perturbations, so it differs from the training-only
fixed basis of \S\ref{sec:methods_audit}. Table~\ref{tab:supp_txpert} reports the
distance score and the projection test. Keeping only the top direction
of the latent lowers mean PDC from $0.599$ to $0.366$ (Exphormer) and
from $0.595$ to $0.344$ (GAT), a random direction lowers it to $0.031$ and
$-0.176$, and eight directions keep $90\%$ of it. Controlling for the magnitude
of the predicted response, the
score retains partial $|\rho| = 0.33$ (Exphormer) and $0.13$ (GAT), both
with intervals excluding zero. The predicted responses have $0.32\times$
(Exphormer) and $0.42\times$ (GAT) the participation ratio of the
observed responses. A released cross-cell-line transfer checkpoint,
trained on other cell lines and evaluated on K562, retains $90.5\%$ of
PDC at rank $16$, with predicted participation ratio $0.29\times$
observed.

\begin{table}[h]
\centering\small
\caption{TxPert: distance score and rank-$k$ projection of the
perturbation latent (mean PDC). Exphormer checkpoints: the distance
rows (top block) use \texttt{K562\_unseen\_pert\_exphormer\_mg}
(STRING and GO graphs); the projection rows (bottom block) and the
participation ratio in the text use
\texttt{K562\_unseen\_pert\_exphormer} (STRING graph).}
\label{tab:supp_txpert}
\begin{tabular}{lcc}
\toprule
 & Exphormer & GAT \\
\midrule
Mean PDC & 0.600 & 0.595 \\
Oracle distance vs PDC, $|\rho|$ & 0.821 & 0.802 \\
Predicted vs observed distance, $\rho$ & 0.607 & 0.443 \\
Predicted-NN-distance vs PDC, $|\rho|$ & 0.697 & 0.481 \\
\midrule
Top-$1$ singular direction & 0.366 & 0.344 \\
Random direction, rank $1$ & 0.031 & $-0.176$ \\
Top-$8$ singular directions & 0.555 & 0.555 \\
Random subspace, rank $8$ & 0.163 & 0.033 \\
Smallest rank retaining $90\%$ & 8 & 8 \\
\bottomrule
\end{tabular}
\end{table}

\subsection{Chemical perturbations in Tahoe-100M}
\label{sec:supp_rev_tahoe}

We used Arc's released \texttt{ST-SE-Tahoe} checkpoint and its
published evaluation predictions on the five few-shot cell lines of
Tahoe-100M~\citep{zhang2025tahoe}; we trained no model. In each of
these cell lines, $402$ drug--dose conditions are in the checkpoint's
training data and $734$--$735$ conditions from $268$ drugs are
evaluated. Responses are
computed as differences of log-normalized means; our PDC reproduces
Arc's published \texttt{pearson\_delta} column at Pearson $1.000000$.
The training pool for the distance is the set of training conditions of
the same cell line. Intervals resample whole drugs so that doses of one
compound are not treated as independent, and permutation tests permute
drugs.

\begin{table}[h]
\centering\small
\caption{Tahoe-100M chemical perturbations: Spearman $\rho$ between
predicted-NN-distance and per-condition PDC with drug-clustered $95\%$
intervals, and participation ratio (PR) of observed and predicted
responses.}
\label{tab:supp_tahoe}
\begin{tabular}{lcccc}
\toprule
Cell line & $\rho$ & $95\%$ interval & PR observed & PR predicted \\
\midrule
C32       & $-0.849$ & $[-0.871, -0.821]$ & 3.48 & 2.39 \\
HOP62     & $-0.788$ & $[-0.819, -0.749]$ & 5.55 & 4.44 \\
HepG2-C3A & $-0.683$ & $[-0.723, -0.635]$ & 5.60 & 4.12 \\
Hs 766T   & $-0.726$ & $[-0.767, -0.680]$ & 7.79 & 5.92 \\
PANC-1    & $-0.710$ & $[-0.750, -0.667]$ & 7.03 & 5.37 \\
\bottomrule
\end{tabular}
\end{table}

The association holds when references to the same drug are removed from
the pool for every condition ($-0.678$ to $-0.845$), on the conditions
whose drug is absent from the pool ($-0.694$ to $-0.821$), when doses are
collapsed to one point per drug ($-0.630$ to $-0.855$), and after
controlling for alignment with the leading response axis (partial
$-0.599$ to $-0.681$). The oracle distance gives $-0.780$ to $-0.874$.

\subsection{Single-cell resolution and context size}
\label{sec:supp_rev_singlecell}

Table~\ref{tab:supp_singlecell_context} reports the association
between the score and per-perturbation PDC, with PDC computed from
pseudobulk or from individual cells, and with the number of context
cells $S$ varied. Across $S \in \{16, 32, 64, 128, 256\}$, every
configuration keeps a negative association with $|\rho| \geq 0.30$,
and every paired-bootstrap interval for the change from $S = 64$ lies
within $\pm 0.10$. The largest point change from the $S = 64$ value is
$0.031$, and the largest range over the five context sizes is $0.048$,
both in Tahoe$+$ST/RPE1. Predictions vary across the cells of a
perturbation, but this variation does not follow the variation between
its observed cells (correlation of the residuals $0.00$--$0.02$). The
single-cell column therefore measures how well the per-perturbation
ranking holds when PDC is computed against individual cells.
Dividing each perturbation's single-cell PDC by its ceiling (the
single-cell PDC of the observed pseudobulk) lowers $|\rho|$ to
$0.04$--$0.47$ for Tahoe$+$ST, while SE$+$ST and scGPT$+$ST keep
$0.55$--$0.73$, so for Tahoe$+$ST part of the single-cell association
reflects the size of the response.

\begin{table}[h]
\centering\small
\setlength{\tabcolsep}{4pt}
\caption{Spearman $|\rho|$ between predicted-NN-distance and PDC:
pseudobulk and single-cell scoring, and pseudobulk scoring at context
size $S$. The $S = 64$ column is an independent rerun of the
pseudobulk column; the two differ by at most $0.020$.}
\label{tab:supp_singlecell_context}
\begin{tabular}{llcccccccc}
\toprule
 & & & & \multicolumn{5}{c}{Context size $S$} \\
\cmidrule(lr){5-9}
Encoder & Cell line & Pseudobulk & Single cell & 16 & 32 & 64 & 128 & 256 \\
\midrule
SE    & HepG2  & 0.844 & 0.744 & 0.852 & 0.841 & 0.850 & 0.837 & 0.841 \\
SE    & Jurkat & 0.832 & 0.647 & 0.862 & 0.852 & 0.849 & 0.855 & 0.858 \\
SE    & K562   & 0.878 & 0.753 & 0.878 & 0.884 & 0.879 & 0.890 & 0.881 \\
SE    & RPE1   & 0.817 & 0.765 & 0.811 & 0.823 & 0.817 & 0.822 & 0.813 \\
Tahoe & HepG2  & 0.840 & 0.712 & 0.870 & 0.855 & 0.841 & 0.833 & 0.839 \\
Tahoe & Jurkat & 0.850 & 0.660 & 0.855 & 0.857 & 0.850 & 0.846 & 0.852 \\
Tahoe & K562   & 0.862 & 0.739 & 0.865 & 0.869 & 0.862 & 0.862 & 0.868 \\
Tahoe & RPE1   & 0.724 & 0.721 & 0.743 & 0.734 & 0.726 & 0.702 & 0.695 \\
scGPT & HepG2  & 0.830 & 0.750 & 0.836 & 0.830 & 0.837 & 0.835 & 0.823 \\
scGPT & Jurkat & 0.852 & 0.724 & 0.852 & 0.844 & 0.832 & 0.847 & 0.841 \\
scGPT & K562   & 0.785 & 0.639 & 0.776 & 0.786 & 0.781 & 0.787 & 0.794 \\
scGPT & RPE1   & 0.559 & 0.566 & 0.550 & 0.533 & 0.559 & 0.561 & 0.556 \\
\bottomrule
\end{tabular}
\end{table}

\subsection{Strict family-held-out retraining}
\label{sec:supp_rev_family_holdout}

We grouped perturbations into eight disjoint gene families (cell
cycle, mitochondrial translation, mitoribosomal, oxidative
phosphorylation, cytoplasmic ribosomal, RNA polymerase, splicing,
translation initiation). For each family and each configuration we
removed every member of the family from training and validation,
retrained the ST decoder from scratch, and computed predicted-NN-distance
and PDC only on that family's held-out perturbations ($96$ retrained
models, plus $12$ ordinary models trained on the same schedule). Within
a configuration we pool the eight held-out families, each predicted by
the model that excluded it. Table~\ref{tab:supp_family_holdout} gives
the pooled Spearman $\rho$. It is negative in $9$ of the $12$
configurations, with joint mean signed $\rho = -0.254$ ($95\%$ interval
$[-0.345, -0.221]$), and a family-stratified centered-rank analysis is
negative in $10$ of $12$. Averaged over leave-one-family tasks, mean
signed $\rho$ is $-0.553$ (HepG2), $-0.416$ (K562), $-0.318$ (Jurkat),
and $+0.071$ (RPE1) by cell line, and $-0.340$ (SE), $-0.301$ (scGPT),
and $-0.271$ (Tahoe) by encoder. The family-clustered $95\%$ intervals
of the three RPE1 configurations include zero. Averaging the Spearman
$\rho$ computed within each family over the twelve configurations, seven
of eight families are negative; some families are small (mitochondrial
translation has seven to nine perturbations per configuration).

\begin{table}[h]
\centering\small
\caption{Strict family-held-out retraining: Spearman $\rho$ between
predicted-NN-distance and PDC on pooled held-out families ($n$
perturbations), and mean PDC.}
\label{tab:supp_family_holdout}
\begin{tabular}{lcccccc}
\toprule
 & \multicolumn{2}{c}{SE$+$ST} & \multicolumn{2}{c}{Tahoe$+$ST} & \multicolumn{2}{c}{scGPT$+$ST} \\
\cmidrule(lr){2-3}\cmidrule(lr){4-5}\cmidrule(lr){6-7}
Cell line ($n$) & $\rho$ & PDC & $\rho$ & PDC & $\rho$ & PDC \\
\midrule
HepG2 ($290$)  & $-0.697$ & 0.251 & $-0.486$ & 0.316 & $-0.588$ & 0.316 \\
Jurkat ($280$) & $-0.363$ & 0.279 & $-0.168$ & 0.281 & $-0.505$ & 0.170 \\
K562 ($313$)   & $-0.474$ & 0.114 & $-0.478$ & 0.176 & $-0.464$ & 0.133 \\
RPE1 ($328$)   & $+0.366$ & 0.471 & $+0.314$ & 0.463 & $+0.497$ & 0.466 \\
\bottomrule
\end{tabular}
\end{table}

\section{Extended related work}
\label{sec:supp_related_extended}

\paragraph{Perturbation prediction before foundation models.}
Latent-arithmetic methods such as scGen~\citep{lotfollahi2019scgen}
treat a perturbation as a vector added in a learned VAE latent space.
Graph models such as GEARS~\citep{roohani2024predicting} use gene
regulatory networks and the Gene Ontology; they work well when the
prior annotation covers a perturbation and less well otherwise.
Optimal-transport methods, including CellOT~\citep{bunne2023learning}
and CINEMA-OT~\citep{dong2023causal}, learn maps between unpaired
control and perturbed cells, and CINEMA-OT separates confounding
variation from the effect of the intervention. Causal representation
learning treats perturbations as sparse mechanism
shifts~\citep{lopez2023learning}, building on identifiability
results~\citep{zhang2023identifiability}.
\citet{dimitrov2026interpretation} review the field. The encoder$+$ST
factorization we study is more recent than these families.

\paragraph{Single-cell foundation models and the encoder--decoder split.}
Geneformer trains a transformer on about $30$M cells with masked rank
prediction~\citep{theodoris2023transfer}, and scGPT uses a generative
masked objective with task-specific fine-tuning~\citep{cui2024scgpt};
both predict perturbations with the same network that encodes the
cell. STATE~\citep{adduri2025predicting} separates the two: a cell-state
encoder (SE-600M) feeds a trainable state-transition decoder (ST).
Tahoe-x1~\citep{tahoex1_2025} is another cell-state encoder, trained on
data that include the Tahoe-100M chemical-perturbation
atlas~\citep{zhang2025tahoe}, and Stack~\citep{dong2026stack} uses
in-context conditioning instead of a separate decoder. We pair the same ST architecture with SE-600M, scGPT, or Tahoe-x1, which separates what the
decoder does from what each encoder's pretraining contributes.

\paragraph{Concurrent work on perturbation prediction.}
VCHarness runs an agentic architecture search over a library of scFM
components~\citep{cheng2026harnessing} on the same four Replogle--Nadig
cell lines. Markov-GPL replaces the count-based pretraining objective
with a Plackett--Luce ranking likelihood~\citep{taggart2026gpl}.
X-Cell is a masked discrete-diffusion transformer with multi-modal
cross-attention~\citep{wang2026xcell}; its weights were not available
in a form we could probe. PerturbDiff~\citep{yuan2026perturbdiff}
treats the perturbed population as a random variable in a reproducing
kernel Hilbert space and runs diffusion over kernel mean embeddings,
and Squidiff~\citep{he2026squidiff} is a conditional diffusion model
in expression space whose semantic latents add for combined
perturbations. These works change the architecture, objective, or
generative model; we study the internal computation of one existing
architecture.

\paragraph{Mechanistic interpretability.}
Activation patching replaces the value of a hidden component with its
value from another run and measures the effect downstream;
\citet{meng2022rome} used it for factual recall in GPT-2, and
\citet{heimersheim2024interp} discuss its use as a method. Causal
tracing of circuits was extended to in-context behavior
by~\citet{wang2022interpretability,olsson2022inductionheads}.
Interventions on low-rank subspaces rely on the linear-feature view
behind steering vectors~\citep{turner2023activation,burns2022discovering}
and sparse autoencoder
dictionaries~\citep{templeton2024scaling,cunningham2023sparse,bricken2023monosemanticity,gao2024toward},
and sparse autoencoders have been applied to protein language
models~\citep{simon2025interplm,parsan2025towards,adams2025mechanistic}.
Our interventions are of these kinds, with the $L_0$ attention output
as the site; replacing the top direction with that of another encoder
or another cell line is the one addition. Causal
abstraction~\citep{geiger2025causal} formalizes the correspondence
between a high-level causal model and the network that implements it.

\paragraph{Perturbation biology and shared responses.}
Genome-scale CRISPRi Perturb-seq~\citep{replogle2022mapping} and its
extension to more cell lines~\citep{nadig2024transcriptome} provide our
benchmark; combinatorial CRISPRa screens~\citep{norman2019exploring}
raise the question of how perturbations compose, which we do not
address. The mean responses in these screens are dominated by a few
co-regulated programs, and \citet{vinastorne2026systema} describe the
shift shared by perturbed cells as systematic variation. We locate this
structure in the decoder ($L_0$, Appendix~\ref{sec:supp_mechanism_extended}),
relate the leading direction to gene programs
(Appendix~\ref{sec:supp_biology}), and show that the direction is
specific to the encoder and the cell line. Benchmarks report that
simple linear baselines match scFM
predictors~\citep{ahlmann2025perturbation,csendes2025benchmarking,wenteler2024perteval},
and \citet{ahlmann2025perturbation} show that PDC rewards predictions
that collapse toward the mean on small-effect perturbations, which
Appendix~\ref{sec:dir4_sigma1_eval} addresses with scores that separate
perturbations.

\section{LLM Usage}
\label{sec:llm_usage}
We used LLM-based assistants, Claude (Anthropic) and Codex (OpenAI), to implement and review analysis code, check analyses, search the literature, and help write and revise the text. We formulated the research questions, designed the study, and interpreted the results, and we take full responsibility for the content of the paper.

\end{document}